\documentclass[10pt,twocolumn,letterpaper]{article}

\usepackage[pagenumbers]{cvpr} 

\usepackage{pgfplots}
\usepackage{graphicx}
\usepackage{amsmath}
\usepackage{multirow} 
\usepackage{amssymb}
\usepackage{booktabs}
\usepackage{color, colortbl}
\usepackage{xspace}
\usepackage{adjustbox}
\usepackage{tablefootnote}
\usepackage{xcolor}
\usepackage[accsupp]{axessibility} 
\definecolor{LightCyan}{rgb}{0.88,1,1}
\definecolor{grey}{rgb}{0.9,0.9,0.9}

\usepackage[pagebackref,breaklinks,colorlinks]{hyperref}

\usepackage[capitalize]{cleveref}
\crefname{section}{Sec.}{Secs.}
\Crefname{section}{Section}{Sections}
\Crefname{table}{Table}{Tables}
\crefname{table}{Tab.}{Tabs.}

\def\cvprPaperID{2842} 
\def\confName{CVPR}
\def\confYear{2024}

\begin{document}

\title{Unknown Prompt, the only Lacuna: Unveiling CLIP's Potential for Open Domain Generalization}

\author{\hspace{2cm}\stepcounter{footnote}Mainak Singha$^{1}\thanks{This work is partially done while studying at IIT Bombay, India}$ \and Ankit Jha$^{2}$ \and Shirsha Bose$^{3}$ \and Ashwin Nair$^{4}$ \hspace{1.8cm} \and \hspace{2.0cm} Moloud Abdar$^{5}$ \and Biplab Banerjee$^{2}$ \hspace{2.0cm} \and \hspace{1.25cm}
$^{1}$ Aisin Corporation, Japan \and 
$^{2}$ IIT Bombay, India \hspace{2.25cm} \and $^{3}$ TU Munich, Germany \and $^{4}$ IISER Thiruvananthapuram, India \and $^{5}$ Deakin University, Australia \and
{\tt\small \hspace{-0.5cm}\{mainaksingha.iitb, ankitjha16, shirshabosecs, ashwin9084yt, m.abdar1987, getbiplab\}@gmail.com}}
\maketitle



\begin{abstract}
We delve into Open Domain Generalization (ODG), marked by domain and category shifts between training's labeled source and testing's unlabeled target domains. Existing solutions to ODG face limitations due to constrained generalizations of traditional CNN backbones and errors in detecting target open samples in the absence of prior knowledge. Addressing these pitfalls, we introduce ODG-CLIP, harnessing the semantic prowess of the vision-language model, CLIP. Our framework brings forth three primary innovations:
Firstly, distinct from prevailing paradigms, we conceptualize ODG as a multi-class classification challenge encompassing both known and novel categories. Central to our approach is modeling a unique prompt tailored for detecting unknown class samples, and to train this, we employ a readily accessible stable diffusion model, elegantly generating proxy images for the open class.
Secondly, aiming for domain-tailored classification (prompt) weights while ensuring a balance of precision and simplicity, we devise a novel visual style-centric prompt learning mechanism.
Finally, we infuse images with class-discriminative knowledge derived from the prompt space to augment the fidelity of CLIP's visual embeddings. We introduce a novel objective to safeguard the continuity of this infused semantic intel across domains, especially for the shared classes.
Through rigorous testing on diverse datasets, covering closed and open-set DG contexts, ODG-CLIP demonstrates clear supremacy, consistently outpacing peers with performance boosts between 8\%-16\%. Code will be available at \url{https://github.com/mainaksingha01/ODG-CLIP}.

\end{abstract}
\begin{figure}
\vspace*{-3.8mm}
    \centering
      \includegraphics[width=1\columnwidth]{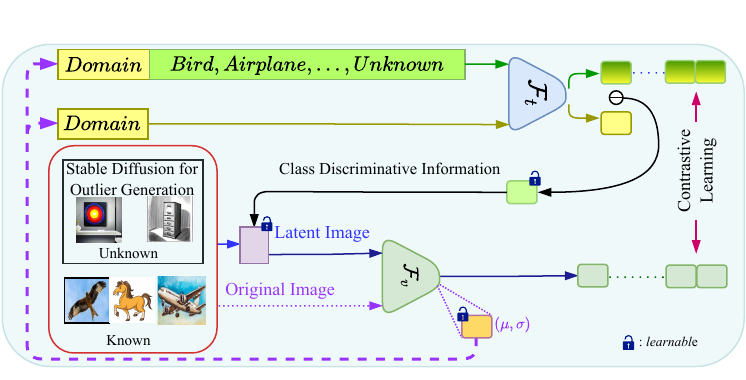}
      \vspace*{-8mm}
    \caption{\textbf{ODG-CLIP} operates as a multi-class classifier leveraging prompt learning for effective management of known categories and outliers in an ODG context. Central to its methodology is a novel \texttt{unknown}-class prompt, designed for open-set samples and integrated with CLIP's unaltered image and text encoders, $\mathcal{F}_v$ and $\mathcal{F}_t$. For the training of \texttt{unknown}-class prompt weights, ODG-CLIP employs pseudo-unknown image generation via stable diffusion (SD)\cite{stablediffusion}. Diverging from existing methods \cite{clip, cocoop, stylip}, ODG-CLIP focuses on creating a refined latent visual space to improve visual embeddings and address domain disparities efficiently. }
    \vspace*{-5.5mm}
    \label{fig:teaser}
\end{figure}

\vspace{-4mm}
\section{Introduction}

Domain Generalization (DG) \cite{wang2022generalizing} outlines an inductive learning strategy wherein a classifier is trained across multiple distinct source domains with diverse data distributions and is then applied to unknown target domains. Unlike closed-set DG, which presupposes uniform semantic categories across domains \cite{li2017deeper, l2a-ot}, Open Domain Generalization (ODG) \cite{odg1} is designed to handle both shared and unique classes within its training ambit. Consequently, during inference, the unlabeled target domain may include both familiar and exclusively new categories. Despite its relevance to various real-world scenarios, such as autonomous driving and remote sensing, research specifically focused on ODG remains scarce.
Prominent ODG techniques, including DAML \cite{odg1} and MEDIC \cite{medic}, incorporate meta-learning to enhance the reliability of classifiers for training classes, thereby facilitating outlier detection alongside recognition of known classes at inference. However, these approaches face obstacles related to the heterogeneous nature of open-domain data. Their generalization capabilities are also compromised, in part because they rely on conventional CNN frameworks, which exhibit limited adaptability.

Foundation models like CLIP \cite{clip} excel by generating rich embedding spaces through multi-modal contrastive training, benefiting especially from prompt learning mechanisms \cite{coop, cocoop} that align classification weights with textual class labels. \textsc{StyLIP} \cite{stylip} advances this by integrating visual style and content in prompt learning, outperforming other counterparts in closed-set DG, yet it has limitations. Initializing prompts from visual properties and refining them through separate projectors adds complexity and may limit the relevance of the downstream target domains. Moreover, \textsc{StyLIP} and similar CLIP-based prompt learning strategies struggle with outlier recognition, reducing their suitability for ODG tasks. Efforts to tackle open-set recognition by enhancing closed-set knowledge \cite{ming2022delving} or developing specialized prompts within the CLIP framework \cite{clipn} have shown promise but remain unoptimized for cross-domain scenarios. The challenge of visual ambiguities also poses significant hurdles for contrastive learning in CLIP, especially across pronounced domain differences.

We identify three key research gaps in using CLIP for ODG:
\textbf{Prompt design}: Emphasizing the superiority of domain conditional prompts in DG \cite{adclip, stylip}, crafting concise, domain-adaptable prompts merging domain-specific and generic tokens is crucial for effective novel domain adaptation.
\textbf{Multi-class classification over one-against-all recourse for ODG \cite{clipn}}: Inspired by the efficacy of a unified multi-class classifier for open-set recognition \cite{morgan, neal2018open}, applying this in a CLIP framework promises enhanced performance but poses challenges, particularly in prompt representation for the open space and sourcing training images.
\textbf{Domain-agnostic visual embeddings}: Boosting the generalizability and discriminative power of visual embeddings could significantly enhance CLIP’s multi-domain efficacy. Investigating the extraction of this information from learned prompts is worth exploring.

\noindent \textbf{Our proposed ODG-CLIP:} We aim to solve the issues in our proposal, ODG-CLIP (Fig. \ref{fig:teaser}), as described below. 

- \textbf{Rethinking CLIP for open sample detection:}  To unify the classification of known classes and outliers using CLIP, we propose a unique \texttt{unknown}-class prompt tailored for detecting the open-set samples. To gather training data for this prompt, our strategy involves generating pseudo-open images that are semantically distinct from existing categories. Moving away from conventional generative or mix-up methods \cite{neal2018open, verma2019manifold}, which often yield semantically inferior images, we opt to leverage a pre-trained conditional diffusion model \cite{stablediffusion}, renowned for its superior and diverse image generation capabilities. To ensure the semantic distinctiveness of these synthesized images, we propose to consider \textit{negative prompts} encompassing the known class names, complementing the \textit{positive prompt} of the form \texttt{Generate [domain] of an unknown class}.

\textbf{- Interplay between novel prompt learning for DG and enhancing visual embeddings from CLIP:} 
Our contributions in this regard encompass two main aspects. Firstly, we introduce an innovative approach to prompt learning for DG that incorporates visual style information from CLIP's vision encoder into a specialized \textit{domain token} while also incorporating semantics through a distinct set of learnable \textit{generic tokens}, demonstrating a better balance between complexity and performance than others counterparts \cite{stylip, maple}.
Secondly, we delineate a technique to augment the caliber of image embeddings. This involves the fusion of a learnable, class-centric channel with the images to create latents that are more discriminative than the raw image data, as is conventionally practised \cite{clip}, for visual feature extraction from CLIP. Precisely, we propose to deploy a dual-prompt strategy per image: one influenced by both style and class information, as aforementioned, and its counterpart driven purely by style. We theorize that the disparities in embeddings of these paired prompts retain class-specific discerning information while capturing the visual distributions of the domain. We introduce a novel loss objective to ensure these differential vectors resonate consistently for communal classes over varied domain pairs.
Our salient contributions are therefore:

 \textbf{[-]} We propose a CLIP-based method, ODG-CLIP, to solve the challenging ODG problem. To our knowledge, ours is the first approach to utilizing vision-language models (VLMs) for solving ODG.

 \textbf{[-]} In ODG-CLIP, we introduce a novel prompt learning for DG, with a specialized prompt to tackle open-class samples and propose a way to use a pre-trained diffusion model to obtain the pseudo-open training samples. Also, we show how the prompt information can be leveraged to enhance the quality of CLIP's visual embeddings.

 \textbf{[-]} We perform extensive validations on open and closed-set DG tasks. ODG-CLIP is found to produce the new state-of-the-art results on six benchmarks for both settings.

\section{Related Works}
\noindent \textbf{Open-set Recognition (OSR) and Open-set Domain Adaptation (OSDA):}
The OSR challenge \cite{osr1,osr2,morgan,osr4} centers on proficiently discerning novel-class samples during evaluation using training exemplars from closed-set classes.
The generative OSR techniques \cite{ge2017generative, du2022vos, zhou2021learning} augment the training set with artificially synthesized categories outside the training set, typically using a GAN-based model or mixing input images randomly to create pseudo-open data \cite{cumix, osr2}. However, these images are mostly restricted in semantics and confined to a lower-dimensional manifold in the open space.
The discriminative models, on the other hand, rely on the confidence of the closed-set classifier, reconstruction loss for the samples, or metric learning to detect the open data \cite{chen2021adversarial, yoshihashi2019classification}. Recently, CLIPN \cite{clipn} has introduced a negative prompt learning approach for OSR using CLIP, outperforming other few existing VLM-based counterparts \cite{ming2022delving,fort2021exploring} significantly.
Notably, these models cannot handle distribution shifts between training and test domains. 
Similarly, OSDA \cite{osda, osda2, osda3, osda4} follows an OSR-like scenario within a \textit{transductive} cross-domain setting.

ODG, with its \textit{inductive} nature and absence of target domain knowledge, presents more formidable challenges than OSR and OSDA. Also, diverging from approaches in OSR, we harness a pre-trained conditional diffusion model, yielding images that represent the open space comprehensively.

\noindent \textbf{(Open) DG:} DG refers to the problem of constructing a supervised learning model that is generalizable across target distributions without the availability of any prior. The initial studies in closed-set DG focused on DA models \cite{li2020domain, wang2021respecting, MMD} due to the disparity in domain distributions. Several DG methods have since been developed, such as self-supervised learning \cite{jigen}, ensemble learning \cite{xu2014exploiting}, and meta-learning \cite{learningtolearn, par, fc, mldg, epi-fcr, rsc}. To address the domain disparity, the concept of domain augmentation \cite{sfa, styleneophile, l2a-ot, mixstyle, zhang2022towards} was introduced, which involves generating pseudo-domains and adding them to the available pool of domains. Subsequently, the notion of ODG was introduced in \cite{odg1}, which is based on domain-augmented meta-learning. 
MEDIC \cite{medic} recently tackled some of the problems of \cite{odg1} and proposed to
consider both domain-wise and class-wise gradient matching  to
learn a balanced decision boundary for the closed-set classes. 
Moving forward, \cite{crossmatch} and \cite{bell} further extended the idea of multi-source ODG to accommodate a single source domain. Unlike these models, we focus on exploring VLMs to solve the ODG task from the perspective of prompt processing and tackle the associated challenges.

\noindent\textbf{VLMs and prompt learning:} The advent of multi-modal foundation models significantly enhances textual and visual integration for image recognition, leveraging BERT \cite{bert} and GPT \cite{gpt} alongside CNN and ViT for content analysis. Key VLMs, such as CLIP \cite{clip} and VisualBERT \cite{visualbert}, initially relied on intricate manual prompts. Prompt learning, emerging to customize these prompts for specific tasks, involves methods like \cite{coop, cocoop, lasp, promptsrc, maple, gopro} to make token embeddings learnable, use projector networks for their evolution, or consider the notion of token sharing between the visual and textual modalities for multi-modal prompting. StyLIP \cite{stylip} uniquely adapts prompt learning for DG, focusing on deriving prompt tokens from visual properties. Further discussions on prompt processing are mentioned in the \texttt{Supplementary}.

Our prompts, blending domain-specific and generic tokens, show enhanced domain adaptability compared to \cite{stylip}. Additionally, we explore the novel approach of improving CLIP's visual embeddings via prompt utilization, without the need for fine-tuning CLIP.

\section{Proposed Methodology}

\begin{figure*}
    \centering
    \vspace*{-0.75cm}
    \includegraphics[width=0.93\linewidth]{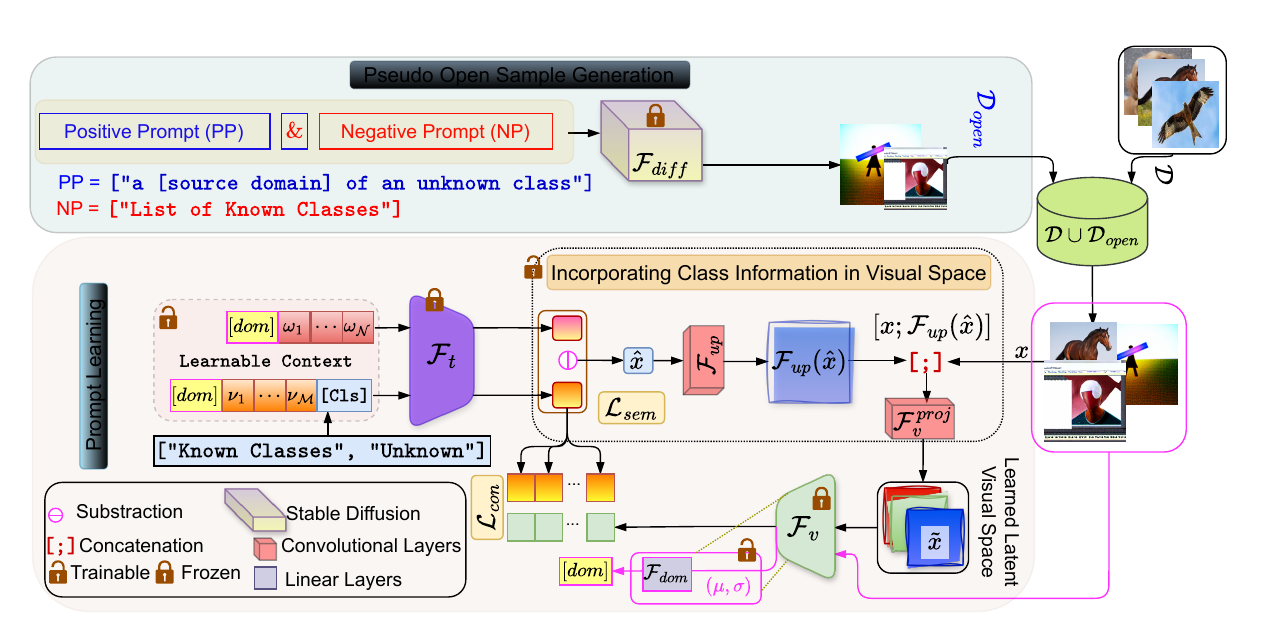}
    \vspace*{-0.5cm}
    \caption{\textbf{Model architecture of ODG-CLIP}, which consists of three main components for designing a multi-class closed-open class classifier using prompt learning with a novel \texttt{unknown}-class prompt for the outliers. Firstly, we propose to generate pseudo-open samples using a pre-trained diffusion model by employing specialized positive and negative textual instructions. The combined images $\mathcal{D} \cup \mathcal{D}_{open}$ go through the prompt learning stage with specialized projectors ($\mathcal{F}_{dom}$), where two types of prompts are learned per image, one using domain+class information, and the other using only domain information. Their difference is used to obtain the latent visual representation $\tilde{x}$ for a given image $x$ conditioned on the class labels from $\mathcal{Y}_{aug}$, through $\mathcal{F}_{up}$ and $\mathcal{F}_v^{proj}$. The model is trained using $\mathcal{L}_{con} + \mathcal{L}_{sem}$ given all the source domains in $\mathcal{D} \cup \mathcal{D}_{open}$. During inference, we create the latent representations for a target image with respect to all the class labels, and the class maximizing Eq. \ref{eq:prob} is selected.}
    \label{fig:model-archi}
    \vspace*{-0.6cm}
\end{figure*}
In the context of ODG, we utilize multiple source domains, denoted as $\mathcal{D} = \{\mathcal{D}_{1}, \mathcal{D}_{2},\cdots,\mathcal{D}_{\mathcal{S}} \}$. Each domain exhibits distinct data distributions and comprises a combination of categories specific to the respective domain and categories shared among them. In the training phase, we employ labeled data samples from each domain $\mathcal{D}_{s} = \{(x_{s}^i,y_{s}^i)\}_{i=1}^{n_s}$, where $y_{s} \in \mathcal{Y}_s$ represents the label associated with $x_{s} \in \mathcal{X}_s$. The total number of unique classes present in $\mathcal{D}$ is denoted as $\mathcal{C}$ for the joint label space $\mathcal{Y}$.

The target domain, denoted as $\mathcal{D}_{\mathcal{T}}=\{x_t^j\}_{j=1}^{n_t}$, differs in its data distribution compared to $\mathcal{D}$. This target domain consists of unlabeled data samples, which can either belong to one of the known classes in $\mathcal{Y}$ or to novel classes not encountered during training. We aim to develop a unified classifier considering the notion of prompt learning in CLIP given $\mathcal{D}$ for effectively discerning outliers while accurately recognizing samples from the known classes in $\mathcal{D}_{\mathcal{T}}$. 

\subsection{Discussing the principles of ODG-CLIP}

\noindent \textbf{Learning objectives:} Our proposed ODG-CLIP is founded on three key principles:
\textbf{i)} Our classification strategy encompasses $\mathcal{C}+1$ classes, where the $\mathcal{C}+1^{th}$ index is designated for the novel \texttt{unknown}-class. This class utilizes specific prompts, the weights of which are shaped by synthetic images generated using a diffusion model \cite{stablediffusion}.
\textbf{ii)} We advocate for adaptive prompt learning across all classes, enabling the capture of domain-specific distributions and overarching semantic contents through distinct token sets.
\textbf{iii)} To further refine visual-textual contrastive learning for ODG-CLIP, we focus on enhancing the discriminability of visual embeddings. This is achieved by establishing a latent visual space, guided by the prompts we've developed. Our goal is to address these aspects cohesively in ODG-CLIP.

\noindent \textbf{Walking through ODG-CLIP:} ODG-CLIP (Fig. \ref{fig:model-archi}) is built upon the frozen vision and text encoders, $\mathcal{F}_v$ and $\mathcal{F}_t$, from CLIP. The pre-trained stable diffusion model $\mathcal{F}_{diff}$ takes a pair of positive and negative prompts (\texttt{PP} and \texttt{NP}) as input, and the generated images are represented by $\mathcal{D}_{open}$. The new training set is $\mathcal{D}_{aug} = \mathcal{D} \cup \mathcal{D}_{open}$, and the label space is $\mathcal{Y}_{aug} = \mathcal{Y} \cup \texttt{unknown}$. In the following, we outline the methodologies implemented to achieve our objectives.

\textbf{- Pseudo-open image synthesis using stable diffusion:}
As mentioned, to train the \texttt{unknown}-class prompts to recognize the outlier samples, we seek to generate pseudo-open samples with improved quality and semantic versatility, which the traditional interpolation/extrapolation or adversarial approaches fail to deliver.
As a remedy, we propose to utilize the pre-trained Stable Diffusion v1-5 model \cite{stablediffusion}. While numerous text-to-image generation techniques exist \cite{dalle, imagen, midjourney}, our preference stems from the impressive inference speed, the latent diffusion models provide. For generating images, we use direct \texttt{PP}  like \texttt{a [domain] of an unknown class} along with \textit{category} \texttt{NP} that include class names from $\mathcal{Y}$, where \texttt{[domain]} refers to the names of the source domains. Our aim is to create potential open-class images that conform to the style characteristics of these known source domains, but diverge from the known semantics of $\mathcal{Y}$ simultaneously.
Positive prompts typically guide the structural composition of images within the latent space. However, by adjusting unconditional sampling to incorporate negative prompts and purposefully excluding negative latents from the conditioned set, we strive to generate images with distinctive characteristics from $\mathcal{D}$.
 Please refer to \cite{stablediffusion} for further details.

However, upon observation, it became evident that some of the generated images were devoid of significant semantics, leading to images mainly marked by uniform regions. To address this issue, we decided against incorporating such subpar images. Instead, we introduced a simple \textit{filtering} method: applying a threshold to the entropy of the grayscale renderings of the generated images, proficiently weeding out the low-entropy images from $\mathcal{D}_{open}$.

\textbf{- Domain-aware prompt learning:} Our method offers a dynamic approach to prompt learning, seamlessly weaving style elements from the visual domain into a set of task-specific, learnable tokens. This form of domain conditioning proves highly effective for DG, allowing the model to adapt fluidly to new target domains as needed. We adopt a straightforward technique to capture domain information, utilizing visual feature statistics from $\mathcal{F}_v$, akin to the method used in \cite{stylip}. However, our approach diverges from \cite{stylip}'s more complex setup, which initializes all tokens with multi-scale visual style or content features. Instead, we integrate a singular domain token, with the remaining tokens positioned as adaptable variables, tailored to suit the task-centric requirements. This configuration achieves a more harmonious balance between task specificity and domain adaptability compared to \cite{stylip}.

Specifically, we propose defining two types of prompts for serving classification and upgrading the visual embeddings simultaneously. One is conditioned on domain and class information, denoted as $\mathcal{P}_{dom_x, cls}(x) = \Bigl[[dom_x], [\nu_1], \cdots, [\nu_{\mathcal{M}}], [cls]\Bigr]$. The other is conditioned solely on the domain, denoted as $\mathcal{P}_{dom_x}(x) = \Bigl[[dom_x], [\omega_1], \cdots, [\omega_{\mathcal{N}}]\Bigr]$. The domain token, $[dom_x]$, is based on the mean and standard deviation, calculated from the final visual feature embedding $\mathcal{F}_v(x)$ for $x$, represented as $[\mu^x; \sigma^x]$. It is mapped to the textual space through the projector $\mathcal{F}_{dom}$: $dom_x = \mathcal{F}_{dom}([\mu^x;\sigma^x])$. $[cls]$ denotes the embeddings of the class names in $\mathcal{Y}_{aug}$.
In contrast, the generic tokens $\nu$ and $\omega$ are directly learned. $\mathcal{F}_t(\mathcal{P}_{dom, cls})$ and $\mathcal{F}_t(\mathcal{P}_{dom})$ denote the prompt embeddings.

\textbf{- Generating latent visual space for obtaining improved visual embeddings:} To enhance visual-textual contrastive learning with $\mathcal{L}_{con}$ (Eq. \ref{eq:1}) amidst domain diversity, we aim to create a semantically rich and class-discriminative latent space from images. We intend to ensure uniformity in features $\mathcal{F}_v$ extracted from these representations for similar-class samples across domains, guided by our consistency loss $\mathcal{L}_{sem}$ (Eq. \ref{eq:2}). Our method leverages prompt embeddings $\mathcal{F}_t(\mathcal{P}_{dom, cls})$ and $\mathcal{F}_t(\mathcal{P}_{dom})$ to imbue images with discriminative qualities, facilitating latent space formation. These prompts inherently capture domain nuances, making the latent space robust against overfitting.
Precisely, our proposed approach involves generating a discriminative knowledge per class, denoted by $\hat{x}^y$, for a given $x$, via element-wise subtraction ($\ominus$) of $\mathcal{F}_t(\mathcal{P}_{dom_x, y}(x))$ from $\mathcal{F}_t(\mathcal{P}_{dom_x}(x))$. An upsampling convolution block, $\mathcal{F}_{up}$, then resizes $\hat{x}^y$ to align with the spatial dimensions of $x$. Following this, $x$ and $\mathcal{F}_{up}(\hat{x}^y)$ are concatenated and processed through the vision projector $\mathcal{F}_{v}^{proj}$, yielding the latent representation $\tilde{x}^y = \mathcal{F}_{v}^{proj}([x;\mathcal{F}_{up}(\hat{x}^y)])$ and which is then inputted into $\mathcal{F}_v$ to extract the visual embeddings $\mathcal{F}_v(\tilde{x}^y)$. This is followed for all $y \in \mathcal{Y}_{aug}$ to aid in the calculation of the loss functions.
\vspace*{-0.05cm}
\subsection{Loss functions, training, and inference }


\noindent \textbf{$\mathcal{L}_{con}$: Visual-textual contrastive learning using proposed embeddings:} We train the prompt tokens using $\mathcal{D}_{aug}$ for both known and unknown class names in $\mathcal{Y}_{aug}$. In opposition to all the CLIP-based models \cite{clip, coop, stylip} that take $\mathcal{F}_v(x)$ as the visual embeddings for contrastive learning, our proposal considers $\mathcal{F}_v(\tilde{x})$ instead.
Precisely, given $\mathcal{F}_v(\tilde{x}^y)$ and $\mathcal{F}_t(\mathcal{P}_{dom_x,y}(x))$, $\mathcal{L}_{con}$ is defined as follows,
\begin{equation}
\begin{split}
    \mathcal{L}_{con} = & \operatorname*{min}_{\mathcal{P}_{dom,class}, \mathcal{P}_{dom}, \mathcal{F}_{dom}, \mathcal{F}_{up}, \mathcal{F}_v^{proj}}  \\ & \underset{(x,y) \in \mathcal{P}(\mathcal{D}_{aug})}{\mathbb{E}}  -  \log p(y | \tilde{x}^y,x)
\end{split}
\label{eq:1}
\end{equation}
\noindent where we calculate $p(y|\tilde{x}^y,x)$ as follows, given $\delta$ as the \texttt{cosine} similarity and $\tau$ as a hyper-parameter.
\vspace*{-0.13cm}
\begin{equation}
\vspace*{-0.05cm}
    p(y|\tilde{x}^y,x) = \frac{\exp(\delta(\mathcal{F}_t(\mathcal{P}_{dom_x,y}(x)), \mathcal{F}_v(\tilde{x}^y))/\tau)}{{\underset{y' \in \mathcal{Y}_{aug}}{\sum}} \exp(\delta(\mathcal{F}_t(\mathcal{P}_{dom_x,y'}(x)), \mathcal{F}_v(\tilde{x}^{y'}))/\tau)}
    \vspace*{-0.1cm}
    \label{eq:prob}
\end{equation}

An intriguing aspect of $\mathcal{L}_{con}$ is that while the generated latent space is utilized for visual feature extraction, the style information is derived from the original image in the prompt embeddings. This intuitive strategy forms the cornerstone of our enhanced contrastive mapping.

 \noindent \textbf{$\mathcal{L}_{sem}$: Proposed cross-domain semantic consistency loss:} Our objective is to ensure uniformity in the information derived from $|\mathcal{F}_t(\mathcal{P}_{dom, cls}) \ominus \mathcal{F}_t(\mathcal{P}_{dom})|$ through $\mathcal{L}_{sem}$, aiming to cultivate a robust class-wise correlation in the derived latent visual representations across images from different domains but sharing identical class labels. Additionally, this helps implicitly disentangle style and semantics within the prompt space, leading to a highly generic embedding space. The other prompt learning approaches, including \cite{stylip}, do not offer such insights.

To illustrate this mathematically, consider two images, $x_i$ and $x_j$, both belonging to class $y$ but originating from different source domains, designated as $(x^i_k, y) \in \mathcal{D}_k$ and $(x^j_l, y) \in \mathcal{D}_l$, respectively, and $\mathcal{D}_k, \mathcal{D}_l \in \mathcal{D}_{aug}$. $\mathcal{L}_{sem}$ is then conceptualized as the \texttt{cosine} distance between their respective prompt differential vectors, as follows,
\vspace*{-0.18cm}
\begin{equation}
\hspace{-.2cm}
\begin{split}
    \centering
    & \mathcal{L}_{sem}   = \operatorname*{min}_{\mathcal{P}_{dom,class}, \mathcal{P}_{dom}} \underset{\mathcal{P}(\mathcal{D}_{aug})}{\mathbb{E}} (1- \delta(|\mathcal{F}_t(\mathcal{P}_{dom_{x_k^i},y}(x_k^i)) \\ & \ominus  \mathcal{F}_t(\mathcal{P}_{dom_{x_k^i}}(x_k^i))|, |\mathcal{F}_t(\mathcal{P}_{dom_{x_l^j},y}(x_l^j)) \ominus \mathcal{F}_t(\mathcal{P}_{dom_{x_l^j}}(x_l^j))|))
\end{split}
\hspace{-.3cm}
\vspace{-0.9cm}
\label{eq:2}
\end{equation}

\noindent \textbf{Total loss:} We train the prompts and the projectors ($\mathcal{F}_{dom}, \mathcal{F}_{up}, \mathcal{F}_v^{proj}$) with the combined loss $\mathcal{L}_{con} + \mathcal{L}_{sem}$.\\
\noindent \textbf{Inference:} During inference, we generate $\tilde{x}_t^{y'}$ for a given $x_t \in \mathcal{D}_{\mathcal{T}}$ for all the $y' \in \mathcal{Y}_{aug}$. The $y'$ maximizing $p(y'|\tilde{x}_t^{y'},x_t)$ (Eq. \ref{eq:prob}) is selected as the predicted label.


\begin{table*}[htbp]
\vspace*{-1.5mm}
\caption{Comparative analysis for PACS, VLCS, Office-Home, Digits-DG, Multi-Dataset and Mini-DomainNet in ODG setting on average ACC and H-score over all the domain combinations following leave-one-domain-out protocol. Here, SD \cite{stablediffusion} represents stable diffusion.}
\vspace*{-6mm}
\begin{center}
\scalebox{0.6607}{
\begin{tabular}{llcccccccccccc|cc}
\toprule

&\multicolumn{1}{l}{\multirow{2}{*}{\textbf{Methods}}}&\multicolumn{2}{c}{\textbf{PACS}} &\multicolumn{2}{c}{\textbf{VLCS}}&\multicolumn{2}{c}{\textbf{OfficeHome}} &\multicolumn{2}{c}{\textbf{Digits-DG}}&\multicolumn{2}{c}{\textbf{Multi-Dataset}}&\multicolumn{2}{c}{\textbf{Mini-DomainNet}}&\multicolumn{2}{|c}{\textbf{Average}}\\
\cmidrule(lr){3-4}\cmidrule(lr){5-6}\cmidrule(lr){7-8}\cmidrule(lr){9-10}\cmidrule(lr){11-12}\cmidrule(lr){13-14}\cmidrule(lr){15-16}
 &&\textbf{Acc}&\textbf{H-score}&
 \textbf{Acc}&\textbf{H-score}&
 \textbf{Acc}&\textbf{H-score}&
 \textbf{Acc}&\textbf{H-score}&
 \textbf{Acc}&\textbf{H-score}&
 \textbf{Acc}&\textbf{H-score}&\textbf{Acc}&\textbf{H-score}\\
\midrule
\multirow{4}{*}{\rotatebox{90}{CNN-based}} &Cumix \cite{cumix} &57.85 &41.05 &52.46 &50.11&51.67& 49.40&58.13& 54.20&42.18 &46.91&50.27&39.16&52.09&46.81\\
&MixStyle \cite{mixstyle}&63.35&48.30&52.30 &50.61&53.52 &49.53&60.23 &56.35&42.18 &46.91&50.43&40.25&53.67&48.66\\
&DAML \cite{odg1}    &65.49 &51.88 &53.53&51.59&56.45 &53.34&59.51&55.61&46.61&51.71&52.81&43.63&55.73&51.29\\

&MEDIC \cite{medic} &89.81 &83.03 &57.28 &55.73 &60.26 &57.91 &\cellcolor{red!20}83.28 &66.30 &50.74 &53.13 &55.29 &45.71 &66.11 &60.30\\
\midrule
\multirow{15}{*}{\rotatebox{90}{CLIP-based}}&\cellcolor{grey}CLIP \cite{clip} &95.16 &76.77 &91.84 &72.94 &81.43 &63.62&77.08&61.95&77.88&72.19&84.50&68.94 &84.65 &69.40\\

&\cellcolor{grey}CLIP + OpenMax \cite{osr1} &93.45 &79.13 &92.09 &\cellcolor{red!20}73.67 &81.00 &61.54 &76.93 &62.78 &78.34 &73.26 &81.89 &69.40 &83.95 &69.96 \\

&\cellcolor{grey}CLIP + OSDA \cite{osda} &92.62 &75.40 &90.21 &70.89 &82.58 &67.35 &80.53 &65.70 &74.45 &75.22 &82.00 &73.62 &83.73 &71.36 \\



&\cellcolor{cyan!10}CoOp \cite{coop} &78.77 &26.87 &92.02 &39.26 &73.85 &36.26&58.54&34.81&66.03	&44.34&61.13&68.34 &71.72&41.65\\

&\cellcolor{cyan!10}CoCoOp \cite{cocoop} &85.76 &32.93 &89.47 &37.01 &75.38 &34.38&52.77&33.50&64.84	&47.57&60.63&56.30&71.48&40.28\\

&\cellcolor{cyan!10}MaPLe \cite{maple} &93.97 &48.47 &89.70 &43.33 &79.47 &33.06 &70.54 &43.83 &69.34	&62.20&74.67&60.57&79.62&48.58\\

&\cellcolor{cyan!10}LASP \cite{lasp} &88.45 &30.37 &90.67 &39.41 &76.13 &34.52 &60.89 &35.23 &66.78 &50.22 &62.34 &61.56 &74.21 &41.89\\

&\cellcolor{cyan!10}PromptSRC \cite{promptsrc} &94.53 &43.32 &90.13 &42.78 &80.21 &36.40 &75.34 &44.25 &65.51 &59.45 &73.60 &62.56 &79.89 &48.13\\

&\cellcolor{cyan!10}CLIPN \cite{clipn} &96.24 &45.00 &84.82 &50.72 &84.55 &42.83 &81.70	&45.56 &77.16 &62.60 &77.38 &66.92 &83.64 &52.27\\

&\cellcolor{cyan!10}\textsc{StyLIP} \cite{stylip} &95.36 &50.74 &90.75 &65.66 &84.73 &60.97 &80.59 &58.15 &\cellcolor{red!20}79.88 &71.99 &80.22 &69.11 &85.26 &62.77\\

&\cellcolor{cyan!10}CLIPN + \textsc{StyLIP} &\cellcolor{red!20}96.37 &64.46 &84.65 &68.02 &83.67 &76.50 &82.14 &59.24 &76.93 &72.15 &\cellcolor{red!20}86.59 &76.18 &85.06 &69.43\\

&\cellcolor{violet!20}MaPLe + SD & 91.47	&82.60	&91.70&	72.67&	85.02&	80.60&	79.92&	65.82&	77.62&	72.83&	83.79&	79.30&	84.92&	75.64\\

&\cellcolor{violet!20}PromptSRC + SD &93.21 &\cellcolor{red!20}87.95 &90.34 &72.62 &84.60 &\cellcolor{red!20}83.31 &80.92 &65.37 &78.44 &77.89 &83.87 &82.95 &85.23 &78.35\\

&\cellcolor{violet!20}\textsc{StyLIP} + SD &91.78 &87.42 &\cellcolor{red!20}92.11 &73.34 &\cellcolor{red!20}85.51 &81.22 &81.45 &\cellcolor{red!20}68.10 &79.05 &\cellcolor{red!20}78.52 &84.12 &\cellcolor{red!20}83.21 &\cellcolor{red!20}85.67&\cellcolor{red!20}78.64\\

\cmidrule(lr){2-16}

&\cellcolor{blue!20}\textbf{\textsc{ODG-CLIP}} &\cellcolor{blue!20}\cellcolor{blue!20}\textbf{99.53} &\cellcolor{blue!20}\textbf{99.70} &\cellcolor{blue!20}\textbf{95.71}&\cellcolor{blue!20}\textbf{86.53} &\cellcolor{blue!20}\textbf{98.32} &\cellcolor{blue!20}\textbf{96.08}&\cellcolor{blue!20}\textbf{91.53}&\cellcolor{blue!20}\textbf{78.27}&\cellcolor{blue!20}\textbf{84.60}&\cellcolor{blue!20}\textbf{90.00}&\cellcolor{blue!20}\textbf{95.68}&\cellcolor{blue!20}\textbf{94.48}&\cellcolor{blue!20}\textbf{94.23$\pm$0.19}&\cellcolor{blue!20}\textbf{90.84$\pm$0.26}\\

\hline
\end{tabular}}
\label{tab_open}
\end{center}
\vspace*{-7.8mm}
\end{table*}

\section{Experimental Evaluations}
\label{sec:Experimenatl_protocol}
\noindent \textbf{Datasets:} We evaluate the efficacy of ODG-CLIP using six benchmark datasets: PACS \cite{li2017deeper}, VLCS \cite{vlcs}, Office-Home \cite{officehome}, Multi-Dataset \cite{odg1}, Digits-DG \cite{l2a-ot}, and the large-scale Mini-DomainNet \cite{domainnet}. Details on the dataset splits are provided in the \texttt{supplementary}.

\noindent \textbf{Architecture details:} In our proposed architecture, $\mathcal{F}_{up}$ is composed of four transpose convolution layers complemented with ReLU activations. For the final layer, we employ bilinear interpolation to ensure a perfect alignment with the input of $\mathcal{F}_v$. Meanwhile, $\mathcal{F}_{v}^{proj}$ incorporates just one convolutional layer, tasked with reducing the input channel count from four to three. On the other hand, $\mathcal{F}_{dom}$ is designed with a single dense layer. We select ViT-B/32 \cite{vit} as the backbone for $\mathcal{F}_{v}$ and the Transformer \cite{transformer} for $\mathcal{F}_{t}$ for all the CLIP-based experiments.

\noindent \textbf{Training and evaluation protocols:} We conduct training over 10 epochs, starting with a warm-up learning rate of 0.01 and using the Adam optimizer \cite{kingma2014adam} in conjunction with a scheduler. The batch size is set to 32 for the PACS, VLCS, Office-Home, and Digits-DG, while for the Multi-dataset and Mini-DomainNet, we use a batch size of 8. To counteract the bias arising from an excess number of unknown labeled images, we limit our generation to only 25\% of the batch size's amount of images for each source domain during training. While generating the synthetic images, a threshold of $0.2$ was fixed for rejecting images with low-entropy values in $\mathcal{D}_{open}$. We consider a consistent context length of four for all the CLIP-based models, following \cite{coop}. To assess our model's performance, we employ two primary metrics in line with the \texttt{leave-one-domain-out} \cite{stylip} protocol where the model is trained in all but one domain which is used during evaluation. We first use the top-1 accuracy (Acc) to gauge the model's effectiveness on closed-set classes. The harmonic mean (H-score) is also calculated to represent performance across closed-set and open-set samples. For closed-set DG, we showcase the top-1 accuracy. The reported results denote the average over three runs.

\subsection{Comparison to the literature}
\vspace*{-0.2cm}
\noindent\textbf{Competitors:} We carried out in-depth comparisons of our ODG-CLIP model against traditional ODG methods, including \cite{cumix, mixstyle, medic, odg1}. Notably, these methods are grounded on conventional CNN backbones like ResNet-18  \cite{resnet} and use a confidence-driven classification for OSR, where a sample with low classification probability for the closed-set classes is marked as an outlier. Furthermore, we conducted exhaustive evaluations against CLIP-based models:
\textbf{Baseline CLIP} \cite{clip}: This method evaluates the OSR prowess by gauging the prediction confidence of target images concerning manually defined prompts for the known classes, exemplified as \texttt{a photo of a [CLS]}.
\textbf{CLIP features paired with OpenMax (OSR)} \cite{osr1}: Here, we combined pre-trained CLIP features with the OpenMax technique to cultivate a joint closed-open set classifier using the amalgamated source domains. \textbf{CLIP features paired with OSDA} \cite{osda}: We combined the pre-trained CLIP with OSDA-BP \cite{osda} for open-set DA, considering a blended source domain and assuming that the target domain is known.
\textbf{Prompt learning techniques}: We assessed a variety of prompt learning strategies, inclusive of \cite{coop, cocoop, maple, stylip, promptsrc}, adopting a confidence-centric open-set prediction approach, similar to the baseline CLIP evaluation mentioned above. 
\textbf{Incorporation of models with an unknown-class prompt (Model + SD)}: For models such as \cite{maple}, \cite{promptsrc} and \cite{stylip}, we enriched them by adding the \texttt{unknown}-class prompt that relies on our considered diffusion-based pseudo-open sample synthesis for training this prompt. For LASP \cite{lasp}, an extra \texttt{unknown}-class was considered for text-to-text contrastive loss during training, and the open-set novel class samples were classified into this unknown class during evaluation.
\textbf{CLIP-based open-set classification (CLIPN)} \cite{clipn}: This method employs CLIP for OSR through the training of a sophisticated encoder for negative-class tokens. Additionally, we integrated the prompt learning of \textsc{StyLIP} \cite{stylip} into the CLIPN architecture, replacing the hand-crafted tokens.
When it comes to closed-set DG, we scrutinized leading non-CLIP methods \cite{cha2021swad,arpit2022ensemble,dandelionnet}, as well as the prompt learning-based approaches. 

\noindent\textbf{Discussions on ODG and closed-set DG performance:} In Table \ref{tab_open}, we compare ODG across six datasets. ODG-CLIP notably surpasses Cumix \cite{cumix}, MixStyle \cite{mixstyle}, DAML \cite{odg1} and MEDIC \cite{medic} in H-score, posting gains of $44.03\%$, $42.18\%$, $39.55\%$ and $30.54\%$, respectively. Remarkably, against the zero-shot CLIP approach, ODG-CLIP exhibits a marked superiority, registering a significant boost of $21.44\%$ in the H-score. When juxtaposed with other prompt learning techniques, ODG-CLIP continues to impress. While integrating explicit \texttt{unknown}-class prompt learning in \cite{promptsrc, stylip, maple} does improve performance relative to their confidence-centric prediction counterparts, they remain eclipsed by ODG-CLIP.  Furthermore, while CLIPN \cite{clipn} manages to outscore several competitors in H-score, ODG-CLIP ultimately trumps CLIPN, benefiting from its embrace of cross-domain learning — a facet absent in CLIPN. In this regard, \textsc{StyLIP}+CLIPN improves the performance of CLIPN substantially but is still poor than ODG-CLIP by $\approx 21 \%$ in H-score.
Finally, \textsc{StyLIP}+SD provides the best result among the competitors, resulting in the average H-score of $78.64\%$. However, it lags ODG-CLIP, which outputs an average H-score of $90.84\%$. 

This improvement in performance can be attributed to enhanced visual feature extraction and our innovative approach to generalizable prompt learning, which fosters a more integrated alignment of image and prompt attributes. Additionally, by simultaneously addressing open and closed-set classification tasks, we enhance the discriminative quality of the embedding space. This leads to a balanced and harmonious performance across both closed and open classes, a synergy often lacking in other models.

\begin{table}[ht!]
\vspace*{-2mm}
\caption{Mean leave-one-domain-out performance on PACS, VLCS, Office-Home, Digits-DG and Mini-DomainNet for DG.}
\vspace*{-6.8mm}
\begin{center}
\scalebox{0.692}{
\begin{tabular}{llccccc|c}
\hline
&\multicolumn{1}{l}{\textbf{ Methods}} & \textbf{ PACS} & \textbf{ VLCS}  & \textbf{ O.H.} & \textbf{ D-DG} &\textbf{M.DNet} & \textbf{Avg.}  \\ 
\midrule
\multirow{3}{*}{\rotatebox{90}{CNN}}&SWAD \cite{cha2021swad} &88.10 &79.10 & 70.60&-&-&79.27 \\
&EoA \cite{arpit2022ensemble} &88.60 &79.10 &72.50 &- &-&80.07\\
&DandelionNet \cite{dandelionnet}&89.20 &81.60 &70.40&-&-&80.40\\
\midrule

\multirow{8}{*}{\rotatebox{90}{CLIP-based}}&CLIP \cite{clip} &94.89 &82.14 &78.40 &64.59 &78.73&79.75 \\
&CoOp \cite{coop} &97.11 &83.34 &81.33 &77.11 &72.30 &82.23\\
&CoCoOp \cite{cocoop} &96.54 &85.02 &81.05 &79.36 &71.51& 82.70\\
&MaPLe \cite{maple} &97.72 &86.75 &83.52 &80.25 &73.87 & 84.42\\
&LASP \cite{lasp} &97.02 &\cellcolor{red!20}87.25 &84.13 &79.92 &70.67 &83.80 \\
&PromptSRC \cite{promptsrc} &98.02 &86.34 &83.89 &\cellcolor{red!20}82.40 &76.10 &85.35 \\
&\textsc{StyLIP} \cite{stylip} &\cellcolor{red!20}98.17 &87.21 &\cellcolor{red!20}85.94 &81.62 &\cellcolor{red!20}80.43 & \cellcolor{red!20}86.67\\
\cmidrule(lr){2-8}
&\cellcolor{blue!20}\textbf{\textsc{ODG-CLIP}} &\cellcolor{blue!20}\textbf{99.83} &\cellcolor{blue!20}\textbf{95.74} &\cellcolor{blue!20}\textbf{96.91} &\cellcolor{blue!20}\textbf{96.38} &\cellcolor{blue!20}\textbf{96.65} &\cellcolor{blue!20}\textbf{97.10}\\
\hline
\end{tabular}}
\label{tab_closed}
\end{center}
\vspace*{-8.5mm}
\end{table}
In line with the trends observed in ODG tasks, ODG-CLIP consistently outperforms all competitors in closed-set DG tasks across all datasets. Closed-set DG represents a special case of ODG, which does not include any category shift among the domains.
As illustrated in Table \ref{tab_closed}, ODG-CLIP showcases its superiority over traditional DG methods based on the conventional CNN backbones like SWAD \cite{cha2021swad}, EoA \cite{arpit2022ensemble}, and DandelionNet \cite{dandelionnet}. Notably, ODG-CLIP outperforms SWAD, EoA, and DandelionNet by an average Acc of $17.83\%$, $17.03\%$, and $16.70\%$, respectively. Furthermore, when compared against prompt learning benchmarks,  ODG-CLIP demonstrates significantly improved performance, outperforming CoOp \cite{coop} by $14.87\%$, CoCoOp \cite{cocoop} by $14.40\%$, MaPLe \cite{maple} by $12.68\%$, and \textsc{StyLIP} \cite{stylip} by $10.43\%$.




\subsection{Ablation analysis \protect \footnote[1]{More analysis on domain alignment, prompts management, qualitative visualizations, model complexity analysis, \etc are mentioned in the \texttt{supplementary}.}}
\noindent\textbf{(i) How do $\mathcal{L}_{sem}$ and $\hat{x}$ help ODG?} While $\tilde{x}$ introduces class-discriminative information to images, $\mathcal{L}_{sem}$ maintains the consistency of this information across various domains. Their roles are interconnected, influencing the ODG performance significantly.
In this context, we begin by assessing the influence of $\mathcal{L}_{sem}$, as detailed in Table \ref{tab_loss_ablation}. Across all datasets, we notice a consistent enhancement in closed-set performance by approximately $3-4\%$ and nearly a $5\%$ surge in H-score compared to the model without $\mathcal{L}_{sem}$. 
\begin{table}[htbp]
\vspace*{-2.5mm}
\caption{Ablation analysis for $\mathcal{L}_{sem}$ and $\hat{x}$ in our proposed ODG-CLIP. Manual $\hat{x}$ refers to the case where $\hat{x}$ is derived from the ready-made embeddings of the class names.}
\vspace*{-6.8mm}
\begin{center}
\scalebox{0.66}{
\begin{tabular}{lcccccccc}
\hline
\multicolumn{1}{l}{\multirow{2}{*}{\textbf{Methods}}}&\multicolumn{2}{c}{\textbf{PACS}} &\multicolumn{2}{c}{\textbf{O.H.}}&\multicolumn{2}{c}{\textbf{M.Data}} &\multicolumn{2}{c}{\textbf{M.DNet}}\\
\cmidrule(lr){2-3}\cmidrule(lr){4-5}\cmidrule(lr){6-7}\cmidrule(lr){8-9}
 &\textbf{Acc}&\textbf{H}&\textbf{Acc}&\textbf{H}&\textbf{Acc}&\textbf{H}&\textbf{Acc}&\textbf{H}\\
 \midrule

w/o $\hat{x}$ and $\mathcal{L}_{sem}$ &90.47 &88.34 &92.21 &87.00 &73.56 &75.73 &87.24 &83.51\\
w/o $\mathcal{L}_{sem}$, with $\hat{x}$ & 94.21 &92.56 &95.67 &91.56 &80.34 &85.32 &91.24 &90.88\\
\midrule
Manual $\hat{x}$ & 93.54 &92.82 &95.31 &91.22 &78.53 &79.26 &90.65 &86.52\\
\midrule
\textbf{Full (ours)} &\textbf{99.53} &\textbf{99.70} &\textbf{98.32} &\textbf{96.08} &\textbf{84.60} &\textbf{90.00} &\textbf{95.68} &\textbf{94.48}\\
\hline
\end{tabular}}
\label{tab_loss_ablation}
\end{center}
\vspace{-6.9mm}
\end{table}

We also examined the impact of $\hat{x}$ by considering a model configuration in which $\hat{x}$ is not combined with $x$. As indicated in Table \ref{tab_loss_ablation}, incorporating $\hat{x}$ leads to marked improvements in performance for both closed and open classes. Furthermore, we investigated an alternative scenario where $\hat{x}$ is derived from the embeddings of class names. In this setup, each class name yields a singular $\hat{x}$, regardless of the domain distinctions.
This approach contrasts with our methodology, where $\hat{x}$ inherently captures the visual space distribution, reflecting domain-specific dynamics. In comparison, this static method of manually deriving $\hat{x}$ does not account for the variability across domains, leading to lesser adaptability. The advantages of our approach are evident in the results, demonstrating a consistent 5-11\% increase in the H-score compared to the manual approach.

\begin{figure}
 \vspace{-1.5mm}
    \centering
    
    \includegraphics[width=1.\columnwidth]{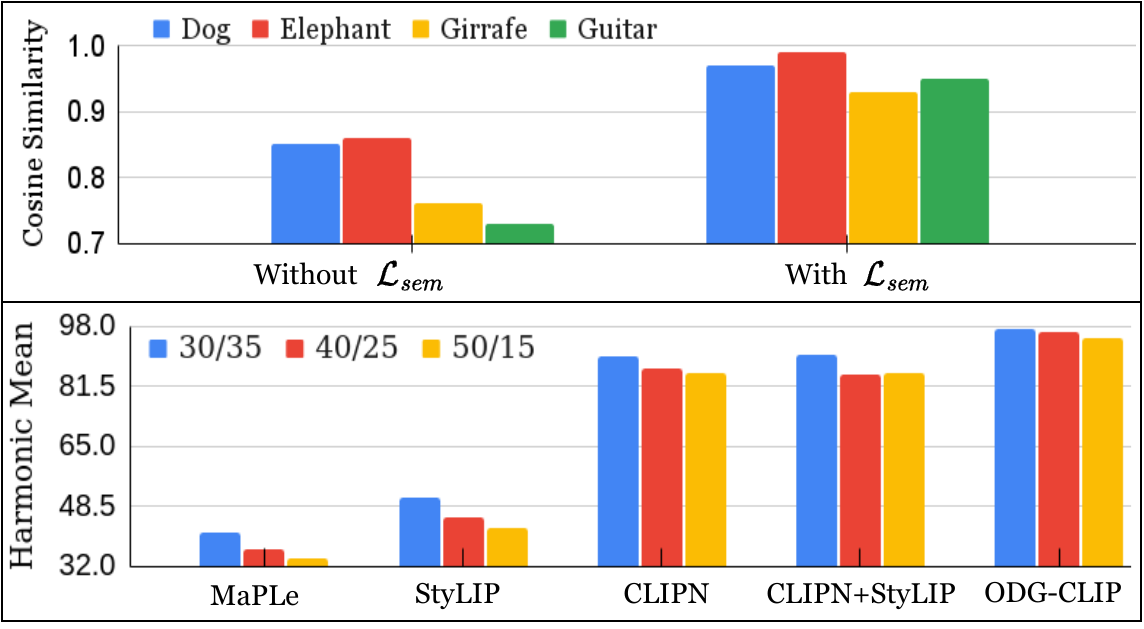}
    \vspace{-8mm}
    \caption{\textbf{Top}: Ablation on the average cosine similarity values of $\hat{x}$ on four shared classes across the domains in PACS, \textbf{Below}: Openness analysis of different methods on Office-Home.}
    
    \label{fig:confusion}
    \vspace{-0.68cm}
\end{figure}

Our hypothesis posits that employing $\mathcal{L}_{sem}$ ensures consistent representation in $\hat{x}$ for images from shared classes across different domains. To substantiate this, we conducted an in-depth analysis of the average pairwise \texttt{cosine} similarity of $\hat{x}$ for four shared classes within the PACS dataset, as depicted in Figure \ref{fig:confusion} (Top). We examined two specific scenarios: \textbf{(i)} where $\hat{x}$ is derived using our method but without the integration of $\mathcal{L}_{sem}$ in our training objective, and \textbf{(ii)} utilizing the complete ODG-CLIP model with $\mathcal{L}_{sem}$.
The results reveal that the full ODG-CLIP model, incorporating $\mathcal{L}_{sem}$, exhibits a higher average \texttt{cosine} similarity compared to the version without $\mathcal{L}_{sem}$. This outcome supports our assertion that $\mathcal{L}_{sem}$ indeed promotes uniformity in the augmented features embedded into the images, reducing the domain divergence considerably. We further report the Fre\'chet \cite{frechet} distance between the source and target domains to justify the better domain alignment offered by ODG-CLIP in the \texttt{Supplementary}.

\noindent\textbf{(ii) Openness analysis of ODG-CLIP:} To assess ODG-CLIP's effectiveness in scenarios with varying levels of 'openness', defined by the number of unknown classes compared to the known classes in $\mathcal{D}_{\mathcal{T}}$, we segmented the Office-Home dataset into subsets with different distributions of known and unknown classes: specifically, splits of $30/35$, $40/25$, and $50/15$. When compared with existing techniques like \cite{maple, stylip, clipn} and their combinations, as shown in Figure \ref{fig:confusion} (Below), ODG-CLIP demonstrates superior capability in differentiating known from unknown classes. This proficiency is highlighted by margins of $7.41\%$, $9.79\%$, and $9.50\%$ in the respective dataset splits.
\begin{figure}[htbp]
    \centering
    \includegraphics[width=0.75\columnwidth]{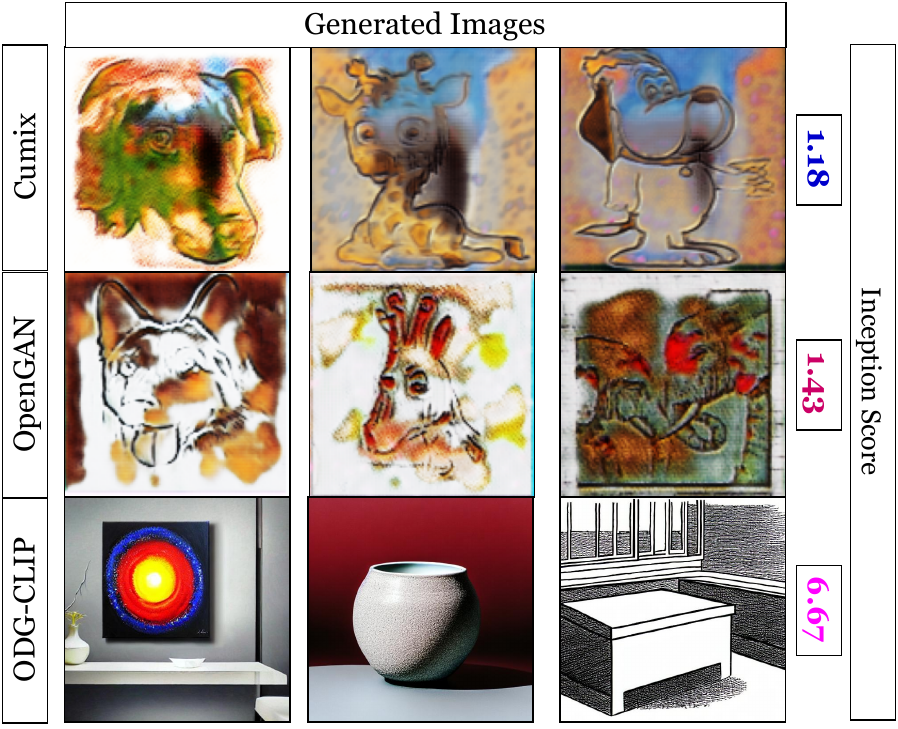}
    \vspace{-3mm}
    \caption{Comparison of ODG-CLIP with Cumix \cite{cumix} and OpenGAN \cite{osr2} on inception score for the generated open-set samples.}
    \vspace{-3mm}
    \label{fig:inception}
\end{figure}

\noindent\textbf{(iii) Comparison of the proposed diffusion-based pseudo-open image generation to the literature:} 
To establish the superiority of the diffusion-based pseudo-open image synthesis, we compared it against two established counterparts. The first, cumix \cite{cumix}, blends two images with a Dirichlet coefficient, classifying the resultant image as outside the known class set. The second, OpenGAN \cite{osr2}, employs adversarial learning to create outlier images distinct from the known classes. We integrated both these methods with ODG-CLIP for open sample synthesis, replacing our diffusion-based approach.

\begin{table}[htbp]
\caption{Analysis of the effects of pseudo-open image generation by our diffusion model and methods from the OSR literature.}
\vspace*{-6mm}
\begin{center}
\scalebox{0.62}{
\begin{tabular}{lcccccccc}
\hline
\multicolumn{1}{l}{\multirow{2}{*}{\textbf{Methods}}}&\multicolumn{2}{c}{\textbf{PACS}} &\multicolumn{2}{c}{\textbf{O.H.}}&\multicolumn{2}{c}{\textbf{M.Data}} &\multicolumn{2}{c}{\textbf{M.DNet}}\\
\cmidrule(lr){2-3}\cmidrule(lr){4-5}\cmidrule(lr){6-7}\cmidrule(lr){8-9}
 &\textbf{Acc}&\textbf{H}&\textbf{Acc}&\textbf{H}&\textbf{Acc}&\textbf{H}&\textbf{Acc}&\textbf{H}\\
\midrule
OpenGAN\cite{osr2} &95.89 &91.35 &91.57 &90.23 &79.93 &80.41 &92.80 &89.76\\
Cumix \cite{cumix} & 95.61 &93.08 &95.42 &92.11 &80.97 &85.92 &91.63 &91.02\\
With SD \cite{stablediffusion} (\textbf{Ours})&\textbf{99.53} &\textbf{99.70} &\textbf{98.32} &\textbf{96.08} &\textbf{84.60} &\textbf{90.00} &\textbf{95.68} &\textbf{94.48}\\
\hline
\end{tabular}}
\label{tab_generation_ablation}
\vspace*{-6mm}
\end{center}
\end{table}

Our analyses, detailed in Table \ref{tab_generation_ablation}, highlight the diffusion method's enhanced performance, evidenced by consistently higher H-scores. This suggests the diffusion model's proficiency in effectively mapping the open space, leading to more accurate classification weights for the \texttt{unknown}-class prompts. The qualitative analysis in Fig. \ref{fig:inception}, along with the inception scores \cite{salimans2016improved} of the generated images, further affirm the diffusion method's effectiveness. Images generated using cumix and OpenGAN exhibited lower quality, as indicated by their inception scores of $1.18$ and $1.35$, respectively. In contrast, the diffusion-based approach successfully synthesized high-quality open-set samples, achieving a significantly higher inception score of $6.67$.

\begin{table}[htbp]
\vspace*{-2mm}
\caption{Analysis of prompts in ODG-CLIP. B1-3 defines the different baseline cases as mentioned below.}
\vspace*{-6.9mm}
\begin{center}
\scalebox{0.68}{
\begin{tabular}{lccccccccl}
\hline
\multicolumn{1}{l}{\multirow{2}{*}{\textbf{Baselines}}}&\multicolumn{2}{c}{\textbf{PACS}} &\multicolumn{2}{c}{\textbf{O.H.}}&\multicolumn{2}{c}{\textbf{M.Data}} &\multicolumn{2}{c}{\textbf{M.DNet}}\\
\cmidrule(lr){2-3}\cmidrule(lr){4-5}\cmidrule(lr){6-7}\cmidrule(lr){8-9}
 &\textbf{Acc}&\textbf{H}&\textbf{Acc}&\textbf{H}&\textbf{Acc}&\textbf{H}&\textbf{Acc}&\textbf{H}\\
\midrule
B1-manual&92.28 &80.56 &84.50 &65.85 &78.23 &75.59 &82.24 &70.41\\
B2-manual&93.42 &84.61 &88.52 &73.89 &80.95 &80.67 &88.56 &83.60\\
B3-Gaussian &93.72 &93.84 &94.60 &88.34 &78.92 &76.34 &90.51 &88.78\\
\midrule
\textbf{ODG-CLIP} &\textbf{99.53} &\textbf{99.70} &\textbf{98.32} &\textbf{96.08} &\textbf{84.60} &\textbf{90.00} &\textbf{95.68} &\textbf{94.48}\\
\hline
\end{tabular}}
\label{tab_prompt_ablation}
\end{center}
\vspace{-7mm}
\end{table}

\noindent\textbf{(iv) Analysis of the prompts in ODG-CLIP:} In order to explore the nuances of prompt processing in ODG-CLIP, we conducted an ablation study, detailed in Table \ref{tab_prompt_ablation}, focusing on three distinct configurations:
\textbf{(B1-manual)}: This setup involves the use of manually defined prompts for both $\mathcal{P}_{dom, cls}$ and $\mathcal{P}_{dom}$, \eg \texttt{[dom] of a [cls]} and \texttt{This is a [dom]}, where [dom] can sketch/painting \etc.
\textbf{(B2-manual)}: In this scenario, the $[dom]$ token is manually initialized, but the generic prompts $(\omega, \nu)$ are set as learnable parameters.
\textbf{(B3-Gaussian)}: This approach aligns with our proposed method for prompt processing, except we initialize ${\omega}$ and ${\nu}$ with noise sampled from a normal distribution $\mathcal{N}(0, \mathbb{I})$. This is in contrast to our full model, which begins the refinement process with these variables initialized from the embedding of the phrase `\texttt{Image of a}'.
The results demonstrate that our strategies for prompt processing outperform the other baselines convincingly.

\noindent\textbf{(v) Impact of \texttt{NP} on performance:} In this regard, we seek to evaluate ODG-CLIP's performance with stable diffusion-driven pseudo-open samples generated using only \texttt{PP} and combined \texttt{PP} and \texttt{NP}. Positive prompts alone proved ambiguous for precise content generation, while negative prompts enhanced clarity on excluding elements for unknown classes. Without \texttt{NP}, the lack of specificity degraded visual-textual mapping in ODG-CLIP, affecting classification accuracy. Employing both positive \texttt{PP} and \texttt{NP} yielded pseudo-open images with varied granularities, improving classifier's robustness for the open samples (Table \ref{tab_generativeprompt_ablation}).
 
\begin{table}[htbp]
\vspace*{-2.5mm}
\caption{Importance of negating prompts (\texttt{NP}) together with positive prompts (\texttt{PP}) for open image synthesis using diffusion.}
\vspace*{-6.5mm}
\begin{center}
\scalebox{0.68}{
\begin{tabular}{lccccccccl}
\hline
\multicolumn{1}{l}{\multirow{2}{*}{\textbf{Methods}}}&\multicolumn{2}{c}{\textbf{PACS}} &\multicolumn{2}{c}{\textbf{O.H.}}&\multicolumn{2}{c}{\textbf{M.Data}} &\multicolumn{2}{c}{\textbf{M.DNet}}\\
\cmidrule(lr){2-3}\cmidrule(lr){4-5}\cmidrule(lr){6-7}\cmidrule(lr){8-9}
 &\textbf{Acc}&\textbf{H}&\textbf{Acc}&\textbf{H}&\textbf{Acc}&\textbf{H}&\textbf{Acc}&\textbf{H}\\
\midrule
Only \texttt{PP} & 92.45 &92.16 &93.72 &90.83 &78.20 &81.57 &91.30 &89.78\\
\texttt{PP} + \texttt{NP} &\textbf{99.53} &\textbf{99.70} &\textbf{98.32} &\textbf{96.08} &\textbf{84.60} &\textbf{90.00} &\textbf{95.68} &\textbf{94.48}\\
\hline
\end{tabular}}
\label{tab_generativeprompt_ablation}
\end{center}
\end{table}



\vspace{-5mm}
\section{Takeaways \& Future Scope}
This paper presents ODG-CLIP, an innovative solution tailored to the complex and relatively unexplored domain of open-domain generalization, viewed through the lens of prompt learning in CLIP. Central to ODG-CLIP are three pivotal innovations: \textit{Domain-aware prompt learning}, \textit{Prompt-driven visual embedding enhancement}, and \textit{Unified classification for known and novel categories}. We develop a unique unknown-class prompt to handle the outliers during testing, specifically trained with data generated by capitalizing on conditional diffusion models.
In our extensive evaluations across both closed-set and open-set DG settings, ODG-CLIP has demonstrated superior performance over existing methodologies. A possible future direction may consider the dense prediction tasks in the ODG setting.

\textcolor{blue}{The authors gratefully \textbf{acknowledge} the immense support provided by Aisin Corporation, Japan.}
{\small
\bibliographystyle{ieee_fullname}
\bibliography{egbib}

\begin{thebibliography}{10}\itemsep=-1pt

\bibitem{arpit2022ensemble}
Devansh Arpit, Huan Wang, Yingbo Zhou, and Caiming Xiong.
\newblock Ensemble of averages: Improving model selection and boosting performance in domain generalization.
\newblock {\em Advances in Neural Information Processing Systems}, 35:8265--8277, 2022.

\bibitem{osr1}
Abhijit Bendale and Terrance~E Boult.
\newblock Towards open set deep networks.
\newblock In {\em Proceedings of the IEEE conference on computer vision and pattern recognition}, pages 1563--1572, 2016.

\bibitem{bommasani2021opportunities}
Rishi Bommasani, Drew~A Hudson, Ehsan Adeli, Russ Altman, Simran Arora, Sydney von Arx, Michael~S Bernstein, Jeannette Bohg, Antoine Bosselut, Emma Brunskill, et~al.
\newblock On the opportunities and risks of foundation models.
\newblock {\em arXiv preprint arXiv:2108.07258}, 2021.

\bibitem{stylip}
Shirsha Bose, Ankit Jha, Enrico Fini, Mainak Singha, Elisa Ricci, and Biplab Banerjee.
\newblock Stylip: Multi-scale style-conditioned prompt learning for clip-based domain generalization.
\newblock In {\em Proceedings of the IEEE/CVF Winter Conference on Applications of Computer Vision}, pages 5542--5552, 2024.

\bibitem{bossard2014food}
Lukas Bossard, Matthieu Guillaumin, and Luc~Van Gool.
\newblock Food-101--mining discriminative components with random forests.
\newblock In {\em European conference on computer vision}, pages 446--461. Springer, 2014.

\bibitem{osda4}
Silvia Bucci, Mohammad~Reza Loghmani, and Tatiana Tommasi.
\newblock On the effectiveness of image rotation for open set domain adaptation.
\newblock In {\em European conference on computer vision}, pages 422--438. Springer, 2020.

\bibitem{lasp}
Adrian Bulat and Georgios Tzimiropoulos.
\newblock Lasp: Text-to-text optimization for language-aware soft prompting of vision \& language models.
\newblock In {\em Proceedings of the IEEE/CVF Conference on Computer Vision and Pattern Recognition}, pages 23232--23241, 2023.

\bibitem{jigen}
Fabio~M Carlucci, Antonio D'Innocente, Silvia Bucci, Barbara Caputo, and Tatiana Tommasi.
\newblock Domain generalization by solving jigsaw puzzles.
\newblock In {\em Proceedings of the IEEE/CVF Conference on Computer Vision and Pattern Recognition}, pages 2229--2238, 2019.

\bibitem{cha2021swad}
Junbum Cha, Sanghyuk Chun, Kyungjae Lee, Han-Cheol Cho, Seunghyun Park, Yunsung Lee, and Sungrae Park.
\newblock Swad: Domain generalization by seeking flat minima.
\newblock {\em Advances in Neural Information Processing Systems}, 34:22405--22418, 2021.

\bibitem{swad}
Junbum Cha, Sanghyuk Chun, Kyungjae Lee, Han-Cheol Cho, Seunghyun Park, Yunsung Lee, and Sungrae Park.
\newblock Swad: Domain generalization by seeking flat minima.
\newblock {\em Advances in Neural Information Processing Systems}, 34:22405--22418, 2021.

\bibitem{chen2021adversarial}
Guangyao Chen, Peixi Peng, Xiangqian Wang, and Yonghong Tian.
\newblock Adversarial reciprocal points learning for open set recognition.
\newblock {\em IEEE Transactions on Pattern Analysis and Machine Intelligence}, 44(11):8065--8081, 2021.

\bibitem{stl}
Adam Coates, Andrew Ng, and Honglak Lee.
\newblock An analysis of single-layer networks in unsupervised feature learning.
\newblock In {\em Proceedings of the fourteenth international conference on artificial intelligence and statistics}, pages 215--223. JMLR Workshop and Conference Proceedings, 2011.

\bibitem{bert}
Jacob Devlin, Ming-Wei Chang, Kenton Lee, and Kristina Toutanova.
\newblock Bert: Pre-training of deep bidirectional transformers for language understanding.
\newblock {\em arXiv preprint arXiv:1810.04805}, 2018.

\bibitem{vit}
Alexey Dosovitskiy, Lucas Beyer, Alexander Kolesnikov, Dirk Weissenborn, Xiaohua Zhai, Thomas Unterthiner, Mostafa Dehghani, Matthias Minderer, Georg Heigold, Sylvain Gelly, et~al.
\newblock An image is worth 16x16 words: Transformers for image recognition at scale.
\newblock {\em arXiv preprint arXiv:2010.11929}, 2020.

\bibitem{frechet}
DC Dowson and BV666017 Landau.
\newblock The fr{\'e}chet distance between multivariate normal distributions.
\newblock {\em Journal of multivariate analysis}, 12(3):450--455, 1982.

\bibitem{du2022vos}
Xuefeng Du, Zhaoning Wang, Mu Cai, and Yixuan Li.
\newblock Vos: Learning what you don't know by virtual outlier synthesis.
\newblock {\em arXiv preprint arXiv:2202.01197}, 2022.

\bibitem{pascalvoc}
Mark Everingham, Luc Van~Gool, Christopher~KI Williams, John Winn, and Andrew Zisserman.
\newblock The pascal visual object classes (voc) challenge.
\newblock {\em International journal of computer vision}, 88(2):303--338, 2010.

\bibitem{vlcs}
Chen Fang, Ye Xu, and Daniel~N. Rockmore.
\newblock Unbiased metric learning: On the utilization of multiple datasets and web images for softening bias.
\newblock In {\em Proceedings of the IEEE International Conference on Computer Vision (ICCV)}, December 2013.

\bibitem{caltech}
Li Fei-Fei, Rob Fergus, and Pietro Perona.
\newblock Learning generative visual models from few training examples: An incremental bayesian approach tested on 101 object categories.
\newblock In {\em 2004 conference on computer vision and pattern recognition workshop}, pages 178--178. IEEE, 2004.

\bibitem{fort2021exploring}
Stanislav Fort, Jie Ren, and Balaji Lakshminarayanan.
\newblock Exploring the limits of out-of-distribution detection.
\newblock {\em Advances in Neural Information Processing Systems}, 34:7068--7081, 2021.

\bibitem{mnist-m}
Yaroslav Ganin and Victor Lempitsky.
\newblock Unsupervised domain adaptation by backpropagation.
\newblock In {\em International conference on machine learning}, pages 1180--1189. PMLR, 2015.

\bibitem{dann}
Yaroslav Ganin, Evgeniya Ustinova, Hana Ajakan, Pascal Germain, Hugo Larochelle, Fran{\c{c}}ois Laviolette, Mario Marchand, and Victor Lempitsky.
\newblock Domain-adversarial training of neural networks.
\newblock {\em The journal of machine learning research}, 17(1):2096--2030, 2016.

\bibitem{clip-adapter}
Peng Gao, Shijie Geng, Renrui Zhang, Teli Ma, Rongyao Fang, Yongfeng Zhang, Hongsheng Li, and Yu Qiao.
\newblock Clip-adapter: Better vision-language models with feature adapters.
\newblock {\em arXiv preprint arXiv:2110.04544}, 2021.

\bibitem{ge2017generative}
ZongYuan Ge, Sergey Demyanov, Zetao Chen, and Rahil Garnavi.
\newblock Generative openmax for multi-class open set classification.
\newblock {\em arXiv preprint arXiv:1707.07418}, 2017.

\bibitem{resnet}
Kaiming He, Xiangyu Zhang, Shaoqing Ren, and Jian Sun.
\newblock Deep residual learning for image recognition.
\newblock In {\em Proceedings of the IEEE conference on computer vision and pattern recognition}, pages 770--778, 2016.

\bibitem{helber2019eurosat}
Patrick Helber, Benjamin Bischke, Andreas Dengel, and Damian Borth.
\newblock Eurosat: A novel dataset and deep learning benchmark for land use and land cover classification.
\newblock {\em IEEE Journal of Selected Topics in Applied Earth Observations and Remote Sensing}, 12(7):2217--2226, 2019.

\bibitem{henaff2020data}
Olivier Henaff.
\newblock Data-efficient image recognition with contrastive predictive coding.
\newblock In {\em International conference on machine learning}, pages 4182--4192. PMLR, 2020.

\bibitem{dandelionnet}
Lanqing Hu, Meina Kan, Shiguang Shan, and Xilin Chen.
\newblock Dandelionnet: Domain composition with instance adaptive classification for domain generalization.
\newblock In {\em Proceedings of the IEEE/CVF International Conference on Computer Vision}, pages 19050--19059, 2023.

\bibitem{rsc}
Zeyi Huang, Haohan Wang, Eric~P Xing, and Dong Huang.
\newblock Self-challenging improves cross-domain generalization.
\newblock In {\em European Conference on Computer Vision}, pages 124--140. Springer, 2020.

\bibitem{huynh2020fine}
Dat Huynh and Ehsan Elhamifar.
\newblock Fine-grained generalized zero-shot learning via dense attribute-based attention.
\newblock In {\em Proceedings of the IEEE/CVF conference on computer vision and pattern recognition}, pages 4483--4493, 2020.

\bibitem{vpt}
Menglin Jia, Luming Tang, Bor-Chun Chen, Claire Cardie, Serge Belongie, Bharath Hariharan, and Ser-Nam Lim.
\newblock Visual prompt tuning.
\newblock In {\em Computer Vision--ECCV 2022: 17th European Conference, Tel Aviv, Israel, October 23--27, 2022, Proceedings, Part XXXIII}, pages 709--727. Springer, 2022.

\bibitem{styleneophile}
Juwon Kang, Sohyun Lee, Namyup Kim, and Suha Kwak.
\newblock Style neophile: Constantly seeking novel styles for domain generalization.
\newblock In {\em Proceedings of the IEEE/CVF Conference on Computer Vision and Pattern Recognition}, pages 7130--7140, 2022.

\bibitem{maple}
Muhammad~Uzair Khattak, Hanoona Rasheed, Muhammad Maaz, Salman Khan, and Fahad~Shahbaz Khan.
\newblock Maple: Multi-modal prompt learning.
\newblock In {\em Proceedings of the IEEE/CVF Conference on Computer Vision and Pattern Recognition (CVPR)}, pages 19113--19122, June 2023.

\bibitem{promptsrc}
Muhammad~Uzair Khattak, Syed~Talal Wasim, Muzammal Naseer, Salman Khan, Ming-Hsuan Yang, and Fahad~Shahbaz Khan.
\newblock Self-regulating prompts: Foundational model adaptation without forgetting.
\newblock In {\em Proceedings of the IEEE/CVF International Conference on Computer Vision}, pages 15190--15200, 2023.

\bibitem{kingma2014adam}
Diederik~P Kingma and Jimmy Ba.
\newblock Adam: A method for stochastic optimization.
\newblock {\em arXiv preprint arXiv:1412.6980}, 2014.

\bibitem{osr2}
Shu Kong and Deva Ramanan.
\newblock Opengan: Open-set recognition via open data generation.
\newblock In {\em Proceedings of the IEEE/CVF International Conference on Computer Vision}, pages 813--822, 2021.

\bibitem{osda3}
Jogendra~Nath Kundu, Naveen Venkat, Ambareesh Revanur, R~Venkatesh Babu, et~al.
\newblock Towards inheritable models for open-set domain adaptation.
\newblock In {\em Proceedings of the IEEE/CVF conference on computer vision and pattern recognition}, pages 12376--12385, 2020.

\bibitem{mnist}
Yann LeCun, L{\'e}on Bottou, Yoshua Bengio, and Patrick Haffner.
\newblock Gradient-based learning applied to document recognition.
\newblock {\em Proceedings of the IEEE}, 86(11):2278--2324, 1998.

\bibitem{lee2018pre}
J~Devlin M Chang~K Lee and K Toutanova.
\newblock Pre-training of deep bidirectional transformers for language understanding.
\newblock {\em arXiv preprint arXiv:1810.04805}, 2018.

\bibitem{li2022learning}
Aodi Li, Liansheng Zhuang, Shuo Fan, and Shafei Wang.
\newblock Learning common and specific visual prompts for domain generalization.
\newblock In {\em Proceedings of the Asian Conference on Computer Vision}, pages 4260--4275, 2022.

\bibitem{mldg}
Da Li, Yongxin Yang, Yi-Zhe Song, and Timothy Hospedales.
\newblock Learning to generalize: Meta-learning for domain generalization.
\newblock In {\em Proceedings of the AAAI conference on artificial intelligence}, volume~32, 2018.

\bibitem{li2017deeper}
Da Li, Yongxin Yang, Yi-Zhe Song, and Timothy~M Hospedales.
\newblock Deeper, broader and artier domain generalization.
\newblock In {\em Proceedings of the IEEE international conference on computer vision}, pages 5542--5550, 2017.

\bibitem{epi-fcr}
Da Li, Jianshu Zhang, Yongxin Yang, Cong Liu, Yi-Zhe Song, and Timothy~M Hospedales.
\newblock Episodic training for domain generalization.
\newblock In {\em Proceedings of the IEEE/CVF International Conference on Computer Vision}, pages 1446--1455, 2019.

\bibitem{li2020domain}
Haoliang Li, YuFei Wang, Renjie Wan, Shiqi Wang, Tie-Qiang Li, and Alex Kot.
\newblock Domain generalization for medical imaging classification with linear-dependency regularization.
\newblock {\em Advances in Neural Information Processing Systems}, 33:3118--3129, 2020.

\bibitem{MMD}
Jingjing Li, Erpeng Chen, Zhengming Ding, Lei Zhu, Ke Lu, and Heng~Tao Shen.
\newblock Maximum density divergence for domain adaptation.
\newblock {\em IEEE Transactions on Pattern Analysis and Machine Intelligence}, 43(11):3918--3930, 2021.

\bibitem{visualbert}
Liunian~Harold Li, Mark Yatskar, Da Yin, Cho-Jui Hsieh, and Kai-Wei Chang.
\newblock Visualbert: A simple and performant baseline for vision and language.
\newblock {\em arXiv preprint arXiv:1908.03557}, 2019.

\bibitem{sfa}
Pan Li, Da Li, Wei Li, Shaogang Gong, Yanwei Fu, and Timothy~M Hospedales.
\newblock A simple feature augmentation for domain generalization.
\newblock In {\em Proceedings of the IEEE/CVF International Conference on Computer Vision}, pages 8886--8895, 2021.

\bibitem{fc}
Yiying Li, Yongxin Yang, Wei Zhou, and Timothy Hospedales.
\newblock Feature-critic networks for heterogeneous domain generalization.
\newblock In {\em International Conference on Machine Learning}, pages 3915--3924. PMLR, 2019.

\bibitem{prod}
Yuning Lu, Jianzhuang Liu, Yonggang Zhang, Yajing Liu, and Xinmei Tian.
\newblock Prompt distribution learning.
\newblock In {\em Proceedings of the IEEE/CVF Conference on Computer Vision and Pattern Recognition}, pages 5206--5215, 2022.

\bibitem{varprompt}
Mohammad Mahdi~Derakhshani, Enrique Sanchez, Adrian Bulat, Victor Guilherme Turrisi~da Costa, Cees~GM Snoek, Georgios Tzimiropoulos, and Brais Martinez.
\newblock Variational prompt tuning improves generalization of vision-language models.
\newblock {\em arXiv e-prints}, pages arXiv--2210, 2022.

\bibitem{cumix}
Massimiliano Mancini, Zeynep Akata, Elisa Ricci, and Barbara Caputo.
\newblock Towards recognizing unseen categories in unseen domains.
\newblock In {\em European Conference on Computer Vision}, pages 466--483. Springer, 2020.

\bibitem{ming2022delving}
Yifei Ming, Ziyang Cai, Jiuxiang Gu, Yiyou Sun, Wei Li, and Yixuan Li.
\newblock Delving into out-of-distribution detection with vision-language representations.
\newblock {\em Advances in Neural Information Processing Systems}, 35:35087--35102, 2022.

\bibitem{neal2018open}
Lawrence Neal, Matthew Olson, Xiaoli Fern, Weng-Keen Wong, and Fuxin Li.
\newblock Open set learning with counterfactual images.
\newblock In {\em Proceedings of the European Conference on Computer Vision (ECCV)}, pages 613--628, 2018.

\bibitem{svhn}
Yuval Netzer, Tao Wang, Adam Coates, Alessandro Bissacco, Baolin Wu, Andrew~Y Ng, et~al.
\newblock Reading digits in natural images with unsupervised feature learning.
\newblock In {\em NIPS workshop on deep learning and unsupervised feature learning}, volume 2011, page~7. Granada, Spain, 2011.

\bibitem{niu2022domain}
Hongjing Niu, Hanting Li, Feng Zhao, and Bin Li.
\newblock Domain-unified prompt representations for source-free domain generalization.
\newblock {\em arXiv preprint arXiv:2209.14926}, 2022.

\bibitem{midjourney}
Jonas Oppenlaender.
\newblock The creativity of text-to-image generation.
\newblock In {\em Proceedings of the 25th International Academic Mindtrek Conference}, pages 192--202, 2022.

\bibitem{morgan}
Debabrata Pal, Shirsha Bose, Biplab Banerjee, and Yogananda Jeppu.
\newblock Morgan: Meta-learning-based few-shot open-set recognition via generative adversarial network.
\newblock In {\em Proceedings of the IEEE/CVF Winter Conference on Applications of Computer Vision (WACV)}, pages 6295--6304, January 2023.

\bibitem{osda}
Pau Panareda~Busto and Juergen Gall.
\newblock Open set domain adaptation.
\newblock In {\em Proceedings of the IEEE international conference on computer vision}, pages 754--763, 2017.

\bibitem{learningtolearn}
Novi Patricia and Barbara Caputo.
\newblock Learning to learn, from transfer learning to domain adaptation: A unifying perspective.
\newblock In {\em Proceedings of the IEEE Conference on Computer Vision and Pattern Recognition}, pages 1442--1449, 2014.

\bibitem{domainnet}
Xingchao Peng, Qinxun Bai, Xide Xia, Zijun Huang, Kate Saenko, and Bo Wang.
\newblock Moment matching for multi-source domain adaptation.
\newblock In {\em Proceedings of the IEEE/CVF international conference on computer vision}, pages 1406--1415, 2019.

\bibitem{visda}
Xingchao Peng, Ben Usman, Neela Kaushik, Judy Hoffman, Dequan Wang, and Kate Saenko.
\newblock Visda: The visual domain adaptation challenge.
\newblock {\em arXiv preprint arXiv:1710.06924}, 2017.

\bibitem{clip}
Alec Radford, Jong~Wook Kim, Chris Hallacy, Aditya Ramesh, Gabriel Goh, Sandhini Agarwal, Girish Sastry, Amanda Askell, Pamela Mishkin, Jack Clark, et~al.
\newblock Learning transferable visual models from natural language supervision.
\newblock In {\em International Conference on Machine Learning}, pages 8748--8763. PMLR, 2021.

\bibitem{gpt}
Alec Radford, Karthik Narasimhan, Tim Salimans, Ilya Sutskever, et~al.
\newblock Improving language understanding by generative pre-training.
\newblock {\em https://www.mikecaptain.com/resources/pdf/GPT-1.pdf}, 2018.

\bibitem{radford2019language}
Alec Radford, Jeffrey Wu, Rewon Child, David Luan, Dario Amodei, Ilya Sutskever, et~al.
\newblock Language models are unsupervised multitask learners.
\newblock {\em OpenAI blog}, 1(8):9, 2019.

\bibitem{dalle}
Aditya Ramesh, Mikhail Pavlov, Gabriel Goh, Scott Gray, Chelsea Voss, Alec Radford, Mark Chen, and Ilya Sutskever.
\newblock Zero-shot text-to-image generation.
\newblock In {\em International Conference on Machine Learning}, pages 8821--8831. PMLR, 2021.

\bibitem{stablediffusion}
Robin Rombach, Andreas Blattmann, Dominik Lorenz, Patrick Esser, and Bj{\"o}rn Ommer.
\newblock High-resolution image synthesis with latent diffusion models.
\newblock In {\em Proceedings of the IEEE/CVF conference on computer vision and pattern recognition}, pages 10684--10695, 2022.

\bibitem{labelme}
Bryan~C Russell, Antonio Torralba, Kevin~P Murphy, and William~T Freeman.
\newblock Labelme: a database and web-based tool for image annotation.
\newblock {\em International journal of computer vision}, 77(1):157--173, 2008.

\bibitem{office31}
Kate Saenko, Brian Kulis, Mario Fritz, and Trevor Darrell.
\newblock Adapting visual category models to new domains.
\newblock In {\em European conference on computer vision}, pages 213--226. Springer, 2010.

\bibitem{imagen}
Chitwan Saharia, William Chan, Saurabh Saxena, Lala Li, Jay Whang, Emily~L Denton, Kamyar Ghasemipour, Raphael Gontijo~Lopes, Burcu Karagol~Ayan, Tim Salimans, et~al.
\newblock Photorealistic text-to-image diffusion models with deep language understanding.
\newblock {\em Advances in Neural Information Processing Systems}, 35:36479--36494, 2022.

\bibitem{osda2}
Kuniaki Saito, Shohei Yamamoto, Yoshitaka Ushiku, and Tatsuya Harada.
\newblock Open set domain adaptation by backpropagation.
\newblock In {\em Proceedings of the European conference on computer vision (ECCV)}, pages 153--168, 2018.

\bibitem{salimans2016improved}
Tim Salimans, Ian Goodfellow, Wojciech Zaremba, Vicki Cheung, Alec Radford, and Xi Chen.
\newblock Improved techniques for training gans.
\newblock {\em Advances in neural information processing systems}, 29, 2016.

\bibitem{tpt}
Manli Shu, Weili Nie, De-An Huang, Zhiding Yu, Tom Goldstein, Anima Anandkumar, and Chaowei Xiao.
\newblock Test-time prompt tuning for zero-shot generalization in vision-language models.
\newblock {\em arXiv preprint arXiv:2209.07511}, 2022.

\bibitem{odg1}
Yang Shu, Zhangjie Cao, Chenyu Wang, Jianmin Wang, and Mingsheng Long.
\newblock Open domain generalization with domain-augmented meta-learning.
\newblock In {\em Proceedings of the IEEE/CVF Conference on Computer Vision and Pattern Recognition}, pages 9624--9633, 2021.

\bibitem{gopro}
Mainak Singha, Ankit Jha, and Biplab Banerjee.
\newblock Gopro: Generate and optimize prompts in clip using self-supervised learning, 2023.

\bibitem{Singha_2023_CVPR}
Mainak Singha, Ankit Jha, Bhupendra Solanki, Shirsha Bose, and Biplab Banerjee.
\newblock Applenet: Visual attention parameterized prompt learning for few-shot remote sensing image generalization using clip.
\newblock In {\em Proceedings of the IEEE/CVF Conference on Computer Vision and Pattern Recognition (CVPR) Workshops}, pages 2023--2033, June 2023.

\bibitem{adclip}
Mainak Singha, Harsh Pal, Ankit Jha, and Biplab Banerjee.
\newblock Ad-clip: Adapting domains in prompt space using clip.
\newblock In {\em Proceedings of the IEEE/CVF International Conference on Computer Vision}, pages 4355--4364, 2023.

\bibitem{singha2023ad}
Mainak Singha, Harsh Pal, Ankit Jha, and Biplab Banerjee.
\newblock Ad-clip: Adapting domains in prompt space using clip.
\newblock {\em arXiv preprint arXiv:2308.05659}, 2023.

\bibitem{transformer}
Ashish Vaswani, Noam Shazeer, Niki Parmar, Jakob Uszkoreit, Llion Jones, Aidan~N Gomez, {\L}ukasz Kaiser, and Illia Polosukhin.
\newblock Attention is all you need.
\newblock {\em Advances in neural information processing systems}, 30, 2017.

\bibitem{osr4}
Sagar Vaze, Kai Han, Andrea Vedaldi, and Andrew Zisserman.
\newblock Open-set recognition: A good closed-set classifier is all you need?
\newblock {\em arXiv preprint arXiv:2110.06207}, 2021.

\bibitem{officehome}
Hemanth Venkateswara, Jose Eusebio, Shayok Chakraborty, and Sethuraman Panchanathan.
\newblock Deep hashing network for unsupervised domain adaptation.
\newblock In {\em Proceedings of the IEEE conference on computer vision and pattern recognition}, pages 5018--5027, 2017.

\bibitem{verma2019manifold}
Vikas Verma, Alex Lamb, Christopher Beckham, Amir Najafi, Ioannis Mitliagkas, David Lopez-Paz, and Yoshua Bengio.
\newblock Manifold mixup: Better representations by interpolating hidden states.
\newblock In {\em International conference on machine learning}, pages 6438--6447. PMLR, 2019.

\bibitem{clipn}
Hualiang Wang, Yi Li, Huifeng Yao, and Xiaomeng Li.
\newblock Clipn for zero-shot ood detection: Teaching clip to say no.
\newblock In {\em Proceedings of the IEEE/CVF International Conference on Computer Vision}, pages 1802--1812, 2023.

\bibitem{wang2022generalizing}
Jindong Wang, Cuiling Lan, Chang Liu, Yidong Ouyang, Tao Qin, Wang Lu, Yiqiang Chen, Wenjun Zeng, and Philip Yu.
\newblock Generalizing to unseen domains: A survey on domain generalization.
\newblock {\em IEEE Transactions on Knowledge and Data Engineering}, 2022.

\bibitem{medic}
Xiran Wang, Jian Zhang, Lei Qi, and Yinghuan Shi.
\newblock Generalizable decision boundaries: Dualistic meta-learning for open set domain generalization.
\newblock In {\em Proceedings of the IEEE/CVF International Conference on Computer Vision}, pages 11564--11573, 2023.

\bibitem{par}
Yufei Wang, Haoliang Li, and Alex~C Kot.
\newblock Heterogeneous domain generalization via domain mixup.
\newblock In {\em ICASSP 2020-2020 IEEE International Conference on Acoustics, Speech and Signal Processing (ICASSP)}, pages 3622--3626. IEEE, 2020.

\bibitem{wang2021respecting}
Ziqi Wang, Marco Loog, and Jan van Gemert.
\newblock Respecting domain relations: Hypothesis invariance for domain generalization.
\newblock In {\em 2020 25th International Conference on Pattern Recognition (ICPR)}, pages 9756--9763. IEEE, 2021.

\bibitem{sundataset}
Jianxiong Xiao, James Hays, Krista~A Ehinger, Aude Oliva, and Antonio Torralba.
\newblock Sun database: Large-scale scene recognition from abbey to zoo.
\newblock In {\em 2010 IEEE computer society conference on computer vision and pattern recognition}, pages 3485--3492. IEEE, 2010.

\bibitem{xu2014exploiting}
Zheng Xu, Wen Li, Li Niu, and Dong Xu.
\newblock Exploiting low-rank structure from latent domains for domain generalization.
\newblock In {\em European Conference on Computer Vision}, pages 628--643. Springer, 2014.

\bibitem{bell}
Shiqi Yang, Yaxing Wang, Kai Wang, Shangling Jui, and Joost van~de Weijer.
\newblock One ring to bring them all: Towards open-set recognition under domain shift.
\newblock {\em arXiv preprint arXiv:2206.03600}, 2022.

\bibitem{yoshihashi2019classification}
Ryota Yoshihashi, Wen Shao, Rei Kawakami, Shaodi You, Makoto Iida, and Takeshi Naemura.
\newblock Classification-reconstruction learning for open-set recognition.
\newblock In {\em Proceedings of the IEEE/CVF Conference on Computer Vision and Pattern Recognition}, pages 4016--4025, 2019.

\bibitem{zhang2022towards}
Hanlin Zhang, Yi-Fan Zhang, Weiyang Liu, Adrian Weller, Bernhard Sch{\"o}lkopf, and Eric~P Xing.
\newblock Towards principled disentanglement for domain generalization.
\newblock In {\em Proceedings of the IEEE/CVF Conference on Computer Vision and Pattern Recognition}, pages 8024--8034, 2022.

\bibitem{amortized}
Xin Zhang, Yusuke Iwasawa, Yutaka Matsuo, and Shixiang~Shane Gu.
\newblock Amortized prompt: Lightweight fine-tuning for clip in domain generalization.
\newblock {\em arXiv preprint arXiv:2111.12853}, 2021.

\bibitem{zhou2021learning}
Da-Wei Zhou, Han-Jia Ye, and De-Chuan Zhan.
\newblock Learning placeholders for open-set recognition.
\newblock In {\em Proceedings of the IEEE/CVF conference on computer vision and pattern recognition}, pages 4401--4410, 2021.

\bibitem{cocoop}
Kaiyang Zhou, Jingkang Yang, Chen~Change Loy, and Ziwei Liu.
\newblock Conditional prompt learning for vision-language models.
\newblock In {\em Proceedings of the IEEE/CVF Conference on Computer Vision and Pattern Recognition}, pages 16816--16825, 2022.

\bibitem{coop}
Kaiyang Zhou, Jingkang Yang, Chen~Change Loy, and Ziwei Liu.
\newblock Learning to prompt for vision-language models.
\newblock {\em International Journal of Computer Vision}, 130(9):2337--2348, 2022.

\bibitem{l2a-ot}
Kaiyang Zhou, Yongxin Yang, Timothy Hospedales, and Tao Xiang.
\newblock Learning to generate novel domains for domain generalization.
\newblock In {\em European conference on computer vision}, pages 561--578. Springer, 2020.

\bibitem{mixstyle}
Kaiyang Zhou, Yongxin Yang, Yu Qiao, and Tao Xiang.
\newblock Domain generalization with mixstyle.
\newblock {\em arXiv preprint arXiv:2104.02008}, 2021.

\bibitem{prograd}
Beier Zhu, Yulei Niu, Yucheng Han, Yue Wu, and Hanwang Zhang.
\newblock Prompt-aligned gradient for prompt tuning.
\newblock {\em arXiv preprint arXiv:2205.14865}, 2022.

\bibitem{crossmatch}
Ronghang Zhu and Sheng Li.
\newblock Crossmatch: Cross-classifier consistency regularization for open-set single domain generalization.
\newblock In {\em International Conference on Learning Representations}, 2021.

\end{thebibliography}
}

\clearpage
\appendix
\section{Contents of the supplementary materials}
In this supplementary document, we present detailed information and further experimental results, including:
\begin{itemize}
\item [1.] \textbf{Dataset Splits for ODG Settings}: Table \ref{split} lists the dataset splits for PACS, VLCS, OfficeHome, DigitDG, Multi-Dataset, and Mini-DomainNet.
\item [2.] \textbf{Extended Literature Survey on Prompt Learning}: An expanded review of prompt learning in CLIP is available in Section \ref{sec:lit}.
\item [3.] \textbf{Implementation Details of Competitors}: Section \ref{sec:imp} elaborates on how competitor models were implemented.
\item [4.] \textbf{Analysis of Fréchet Distance}: In Table \ref{fid}, we analyze the Fréchet distance \cite{frechet} between each source and target domain in the PACS dataset to evaluate domain alignment.
\item [5.] \textbf{Model Complexity Comparison (GFLOPS)}: Figure \ref{fig:gflop} compares different models based on their GFLOPS calculation during training.
\item [6.] \textbf{Ablation Studies}: These include an examination of the domain token position in prompts (Table \ref{dom_token_position}), context length for prompts (Table \ref{Context_length}), and cosine-similarity of $\hat{x}$ features for pseudo-unknown-class samples across domains (Table \ref{tab_cos}).
\item [7.] \textbf{Qualitative Analysis}: Figure \ref{fig:np} highlights the effect of utilizing negative prompts for creating pseudo-open images. Additionally, Figure \ref{fig:tsne1} presents a t-SNE visualization, contrasting our method's latent visual space representation with the traditional hand-crafted $\hat{x}$ for class embeddings. Furthermore, Figure \ref{fig:tsne2} offers a comparative analysis of open samples generated by Cumix \cite{cumix}, OpenGAN \cite{osr2}, and our diffusion model \cite{stablediffusion} within the embedding space.

\item [8.] \textbf{Model Ablation Results}: Table \ref{tab_encoder} shows results for ODG-CLIP using ViT/B-16 and ResNet-50-based CLIP visual encoders.
\item [9.] \textbf{Extended Results with Unknown-Class Prompts}: Table \ref{tab_abl1} extends the (model+SD) results from Table 1 in the main paper.
\item [10.] \textbf{ODG Results on Full DomainNet}: Table \ref{tab_DN} provides detailed results and comparisons for the full DomainNet dataset \cite{domainnet}.
\item [11.] \textbf{Individual Domain Combination Results}: Detailed results for individual domain combinations of open and closed-set DG, supplementing Tables 1 and 2 in the main paper, are presented in Tables \ref{tab_pacs_open} through \ref{tab_digitsmini_closed}.
\end{itemize}

\section{Datasets descriptions}

\textbf{Office-Home} Dataset \cite{officehome}: Comprising 15,500 images, this dataset is divided into 65 classes across four domains: Art, Clipart, Product, and Real.
\textbf{PACS} Dataset \cite{li2017deeper}: The PACS dataset includes 9,991 images, categorized into seven classes and spread over four domains: Artpaint, Cartoon, Sketch, and Photo.
\textbf{VLCS} Dataset \cite{vlcs}: This dataset amalgamates images from four classification datasets (PASCAL VOC 2007 \cite{pascalvoc}, Caltech \cite{caltech}, LabelMe \cite{labelme}, Sun \cite{sundataset}) and consists of images across five classes: Bird, Car, Chair, Dog, and Person.
\textbf{Digits-DG} Dataset \cite{l2a-ot}: Digits-DG is an aggregation of several handwritten digit recognition datasets, including MNIST \cite{mnist}, MNIST-M \cite{mnist-m}, SVHN \cite{svhn}, and SYN \cite{mnist-m}.
\textbf{Multi-dataset} \cite{odg1}: This dataset combines various public datasets such as Office-31 \cite{office31}, STL-10 \cite{stl}, and Visda2017 \cite{visda}, including four domains from DomainNet \cite{domainnet}. It features 20 open classes not present in the source domains' joint label set.
\textbf{Mini-domainnet} \cite{domainnet}: This dataset features four domains, each comprising images from 125 categories.
\textbf{Domainnet} \cite{domainnet}: Comprising six domains, this dataset includes images from 345 categories.
The class splits for all five datasets used in ODG are detailed in Table \ref{class_split}, with classes arranged in alphabetical order.
\begin{table*}[htbp!]
\caption{Dataset splits for the ODG settings: PACS, VLCS, OfficeHome (O.H.), DigitDG (D-DG), Multi-dataset(M.Data), Mini-DomainNet (M.DNet) and DomainNet datasets.}
\vspace*{-3mm}
\begin{center}
\begin{adjustbox}{width=\linewidth}
\begin{tabular}{lccccccc}\toprule
\multicolumn{1}{l}{\textbf{Domain}} &\textbf{PACS} &\textbf{VLCS} &\textbf{OfficeHome} &\textbf{Digits-DG} &\textbf{Multi-Datasets} &\textbf{Mini-DomainNet} &\textbf{DomainNet} \\ 
\midrule
Source 1 & 3, 0, 1 & 0, 1 & 0 - 14,21 - 31 & 0, 1, 2 & 0 - 30 & 0 - 19, 40 - 59 & 0 - 19, 30 - 59, 70 - 99 \\ \hline
& & & & & & & \\
Source 2 & 4, 0, 2 & 1, 2 & 0 - 8, 15 - 20, & 2, 3, 4 & 1, 31 - 41 & 0 - 9, 20 - 39, & 10 - 49, 90 - 129 \\ \hline
& & & 32 - 42 & & & 80 - 89 & \\
Source 3 & 5, 1, 2 & 2, 3 & 0 - 2, 9 - 20, & 4, 5, 6 & 31, 33, 34, & 10 - 19, 40 - 49, & 60 - 79, 140 - 164 \\ 
& & & 43 - 53 & & 41 - 47 & 60 - 79 & 180 - 194, 210 - 229 \\ \hline
Source 4 & - & - & - & - &  - & - & 130 - 139, 160 - 184, \\
& & & & & & & 195 - 219, 250 - 269 \\ \hline
Source 5 & - & - & - & - & - & - & 20 - 39, 220 - 249, 270 - 299 \\
\\
\midrule
& & & 0, 3 - 4, 9 - 10, & & 0, 1, 5, 6, 10, 11, &0 - 4, 8- 17, & 0 - 9, 70 - 79 \\
Target & 0-6 & 0-4 & 15 - 16, 21 - 23, & 0-9 & 14, 17, 20, 26, & 25 - 34, 43 - 47, & 120 - 129, 180 - 189 \\
& & & 32 - 34, 43 - 45, & & 31 - 36, 39 - 43, & 75 - 79, 83 - 87, & 230 - 239, 280 - 289 \\
& & & 54 - 64 & & 45 - 46, 48 - 67 & 90 - 125 & 300 - 344 \\
\bottomrule
\end{tabular}
\end{adjustbox}
\label{class_split}
\end{center}
\label{split}
\end{table*}

\section{Extended literature survey of prompt learning using CLIP}\label{sec:lit}

Vision-Language Models (VLMs) have garnered significant interest across language processing and computer vision fields \cite{lee2018pre, radford2019language, bommasani2021opportunities, bossard2014food, helber2019eurosat,singha2023ad,Singha_2023_CVPR}. These models typically employ task-specific textual descriptions to interpret and analyze visual data \cite{henaff2020data, huynh2020fine}. While early prompting strategies relied on manual definitions, more recent developments have shifted towards automated prompt learning.
CoOp \cite{coop} introduces an approach to optimize both unified and class-specific prompts via back-propagation. CoCoOp \cite{cocoop} further expands on CoOp by incorporating input-conditioned prompt learning, thus addressing issues related to generalization. The CLIP-adapter \cite{clip-adapter} innovates by fine-tuning feature adapters within both the visual and language branches of the model.
ProGrad \cite{prograd} is designed to prevent the forgetting of foundational knowledge within these models. TPT \cite{tpt} leverages the consistency between multiple views of an image for supervision. Probabilistic and variational models such as Prod \cite{prod} and Varprompt \cite{varprompt} focus on learning prompt distributions that align with the spread of visual features. LASP \cite{lasp} enhances the quality of learned prompts through a text-to-text cross-entropy loss. Meanwhile, MaPLe \cite{maple} works on improving the compatibility between different levels of CLIP encoders.
However, a notable limitation of these approaches is their lack of specialization in handling multi-domain data, a crucial aspect for broader applicability in diverse real-world scenarios.

In the realm of domain generalization, several researchers have investigated the concept of domain invariant prompts. For instance, \cite{niu2022domain} and \cite{li2022learning} focus on harnessing text-based source domain knowledge or utilizing image patches as prompt inputs in Vision Transformer (ViT) models. This approach is akin to the methodology used in VPT \cite{vpt}, where prompts are adapted based on specific image features, aiming to achieve a more domain-agnostic model performance. DPL \cite{amortized} employs CLIP \cite{clip} for multi-source Domain Generalization (DG) by deducing domain information from visual features on a batch-wise basis. However, DPL does not fully exploit CLIP’s capability to discern domain-specific details. Additionally, it is prone to overfitting when dealing with small batches, as accurately estimating unbiased style characteristics becomes challenging. 

As can be observed, our prompt learning technique stands out from all the previous literature.

\section{Additional implementation details of the competitor models}\label{sec:imp}

In the CLIP+OpenMax configuration, we have developed a $\mathcal{C}+1$-class, threshold-free classifier using CLIP features to form a unified classifier. For the CLIP+OSDA variant, we incorporate a trainable linear layer on top of the pre-trained CLIP features, which acts as the generator. This is complemented by distinct discriminators for both source-specific classification and domain alignment. The adversarial aspect of this setup is implemented through a gradient-reversal layer, following the methodology outlined in \cite{dann}.

Regarding other prompt learning techniques, our implementation is faithful to the procedures described in the original works. For the CLIPN+\textsc{StyLIP} model, we divide the tokens into two separate categories. One category is shaped by the token learning strategy of \textsc{StyLIP}, and the other consists of specialized tokens that are modified in line with CLIPN's framework. This bifurcated token strategy effectively combines the strengths of both \textsc{StyLIP} and CLIPN, ensuring a harmonious and potent integration of these methodologies.

\section{Analysis of domain alignment using the Fr\'echet distance \cite{frechet} }

Table \ref{frechet_dist} presents the source-to-target domain alignment in various PACS dataset combinations, using the Fréchet distance as a metric. A lower Fréchet distance denotes better domain alignment. In these evaluations, ODG-CLIP demonstrates significant superiority over two main competitors: DAML \cite{odg1}, employing a traditional CNN backbone, and the combined model of CLIPN + \textsc{StyLIP}, using baseline CLIP \cite{clip} features. This advantage of ODG-CLIP is evidenced by its smaller Fréchet distances, indicating more effective domain alignment. Additionally, the impact of excluding the consistency loss $\mathcal{L}_{sem}$ from ODG-CLIP is shown, revealing a decrease in alignment quality compared to the complete ODG-CLIP model.
\begin{table*}[htbp!]
\caption{Ablation study on Fr\'echet distance between each of the source and target domains on PACS dataset using the visual features for domain alignment.}
\vspace*{-3mm}
\begin{center}
\begin{adjustbox}{width=\linewidth}
\begin{tabular}{lccc|ccc|ccc|ccc}\toprule
\multicolumn{1}{l}{\textbf{ Methods}} & \textbf{Cr$\rightarrow$Ar}  &\textbf{Ph$\rightarrow$Ar} &\textbf{Sk$\rightarrow$Ar} & \textbf{Ar$\rightarrow$Cr}  & \textbf{Ph$\rightarrow$Cr} & \textbf{Sk$\rightarrow$Cr} &\textbf{Ar$\rightarrow$Ph} &\textbf{Cr$\rightarrow$Ph} &\textbf{Sk$\rightarrow$Ph} &\textbf{Ar$\rightarrow$Sk} &\textbf{Cr$\rightarrow$Sk} &\textbf{Ph$\rightarrow$Sk} \\ 
\midrule
DAML\cite{odg1} &256.41 &278.35 &224.13 &235.89 &240.14 &197.34 &301.56 &296.31 &283.27 &200.37 &178.92 &235.28 \\ 
CLIP \cite{clip} &231.43 &217.75 &230.32 &224.51 &234.17 &207.21 &267.56 &275.32 &258.48 &160.31 &180.46 &218.35 \\
CLIPN\cite{clipn} + StyLIP\cite{stylip} &200.67 &195.70 &180.35 &198.21 &204.21 &180.25 &247.89 &263.19 &240.38 &149.39 &160.86 &198.37 \\ 
\midrule

\textbf{\textsc{ODG-CLIP}} w/o $\mathcal{L}_{sem}$ &140.22 &135.68 &120.75 &105.43 &145.90 &125.22 &187.33 &189.45 &178.88 &121.22 &142.67 &150.40\\
\textbf{\textsc{ODG-CLIP}} &\textbf{112.56} &\textbf{120.48} &\textbf{95.26} &\textbf{87.32} &\textbf{103.78} &\textbf{105.47} &\textbf{140.26} &\textbf{132.58} &\textbf{146.52} &\textbf{105.37} &\textbf{124.50} &\textbf{131.41}\\
\bottomrule
\end{tabular}
\end{adjustbox}
\label{frechet_dist}
\end{center}
\label{fid}
\end{table*}

\section{Comparison of model complexity for different CLIP based techniques for ODG}

\begin{figure}[ht!]
    \centering
    \includegraphics[width=0.9\linewidth]{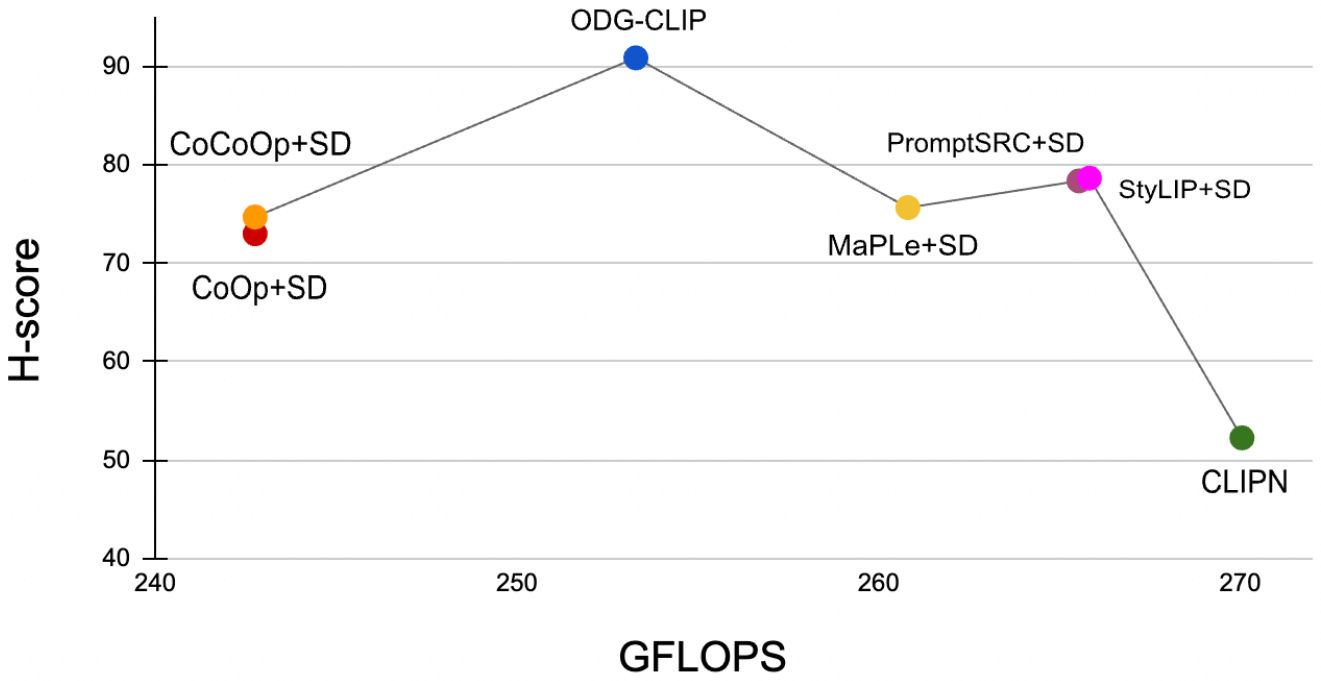}
    \vspace*{-0.5cm}
    \caption{GFLOPs comparison of different methods.}
    \label{fig:gflop}
    \vspace*{0.5cm}
\end{figure}

In Fig. \ref{fig:gflop}, we present a comparison of the model complexity of ODG-CLIP with its competitors. ODG-CLIP exhibits a level of complexity that is on par with most other models, yet it notably surpasses more complex alternatives like \textsc{StyLIP} + SD or CLIPN by a considerable margin. Importantly, when it comes to the H-Score, a key metric of performance, ODG-CLIP consistently outperforms all its counterparts, demonstrating its efficacy despite having comparable complexity.


\section{Additional ablation studies}

\noindent \textbf{Position of the $dom$ token in the prompts}: In Table \ref{dom_token_position}, we present an ablation study that varies the position of domain tokens in $\mathcal{P}_{dom, class}$ and $\mathcal{P}_{dom}$, demonstrated across four datasets. 

\begin{table}[htbp!]
\caption{Ablation on the position of the domain tokens in the prompts.}
\vspace*{-3mm}
\begin{center}
\begin{adjustbox}{width=\linewidth}
\begin{tabular}{cccccccccc}\toprule
{position}&\multicolumn{2}{c}{PACS} &\multicolumn{2}{c}{O.H.}&\multicolumn{2}{c}{M.Data} &\multicolumn{2}{c}{M.DNet}\\
\cmidrule(lr){2-3}\cmidrule(lr){4-5}\cmidrule(lr){6-7}\cmidrule(lr){8-9}
 &Acc &H &Acc &H &Acc &H &Acc &H\\
\midrule
\textit{front} &\textbf{99.53} &\textbf{99.70} &\textbf{98.32} &\textbf{96.08} &84.60 &90.00 &\textbf{95.68} &\textbf{94.48}\\
\textit{middle} &98.40 &98.35 &98.15 &\textbf{96.08} &\textbf{84.63} &\textbf{90.08} &95.51 &93.87\\
\textit{end} &\textbf{99.53} &\textbf{99.70} &98.27&\textbf{96.08} &\textbf{84.63} &90.00 &\textbf{95.68} &\textbf{94.48}\\
\bottomrule
\end{tabular}
\end{adjustbox}
\label{dom_token_position}
\end{center}
\end{table}

\noindent \textbf{Sensitivity of ODG-CLIP to the context lengths of the prompts}:
Table \ref{Context_length} illustrates how ODG-CLIP's performance is affected by the context length in both $\mathcal{P}_{dom, cls}$ and $\mathcal{P}_{dom}$. Generally, a context length of four yields the best outcomes, though a length of 16 also shows comparable results in most cases.

\begin{table}[htbp!]
\caption{Ablation on context lengths. ($\mathcal{M}$, $\mathcal{N}$) depicts the context length of $\mathcal{P}_{dom, cls}$ and $\mathcal{P}_{dom}$. We consider the case when \texttt{Art} serves as the target domain in Office-Home.}
\vspace*{-3mm}
\begin{center}
\begin{adjustbox}{width=\linewidth}
\begin{tabular}{ccccccccc}\toprule
\multicolumn{1}{l}{token length} &(4,4) &(4,28)  &(8,24) &(12,20) &(16,16) &(20,12) &(24,8) & (28,4) \\ 
\midrule
H-score &\textbf{95.88} &93.78 &94.80 &94.80 &\textbf{95.88} &92.83 &92.81 &91.81 \\
\bottomrule
\end{tabular}
\end{adjustbox}
\label{Context_length}
\end{center}
\end{table}

\noindent \textbf{Cosine similarity measurements of latent features $\hat{x}$ for pseudo-unknown class images}:
Building on the findings presented in Fig. 3 (Top) of the main paper, where we explored the impact of $\mathcal{L}_{sem}$ on the cosine similarity of the $\hat{x}$ tensor for closed classes, Table \ref{tab_cos} extends this analysis by demonstrating the effects of $\mathcal{L}_{sem}$ on the $\hat{x}$ information for pseudo-unknown images.

\begin{table}[ht]
\centering
\caption{Cosine similarity in terms of $\hat{x}$ features with and without $\mathcal{L}_{sem}$ for the unknown-class samples averaged over all the domains.}
\label{tab:performance_metrics}
\begin{adjustbox}{width=\linewidth}
\begin{tabular}{|l|c|c|c|c|c|}
\hline
\textbf{Configuration} & \textbf{PACS} & \textbf{VLCS} & \textbf{Office-Home} & \textbf{M-Dataset} & \textbf{M-DomainNet} \\ \hline
With $\mathcal{L}_{sem}$                   & 0.81          & 0.82          & 0.76                & 0.78               & 0.79                \\ \hline
Without $\mathcal{L}_{sem}$               & 0.31          & 0.30          & 0.32                & 0.37               & 0.35                \\ \hline
\end{tabular}
\end{adjustbox}
\label{tab_cos}
\vspace*{-2mm}
\end{table}




\section{Qualitative analysis}

\begin{figure}[htbp!]
    \centering
    \includegraphics[width=\linewidth]{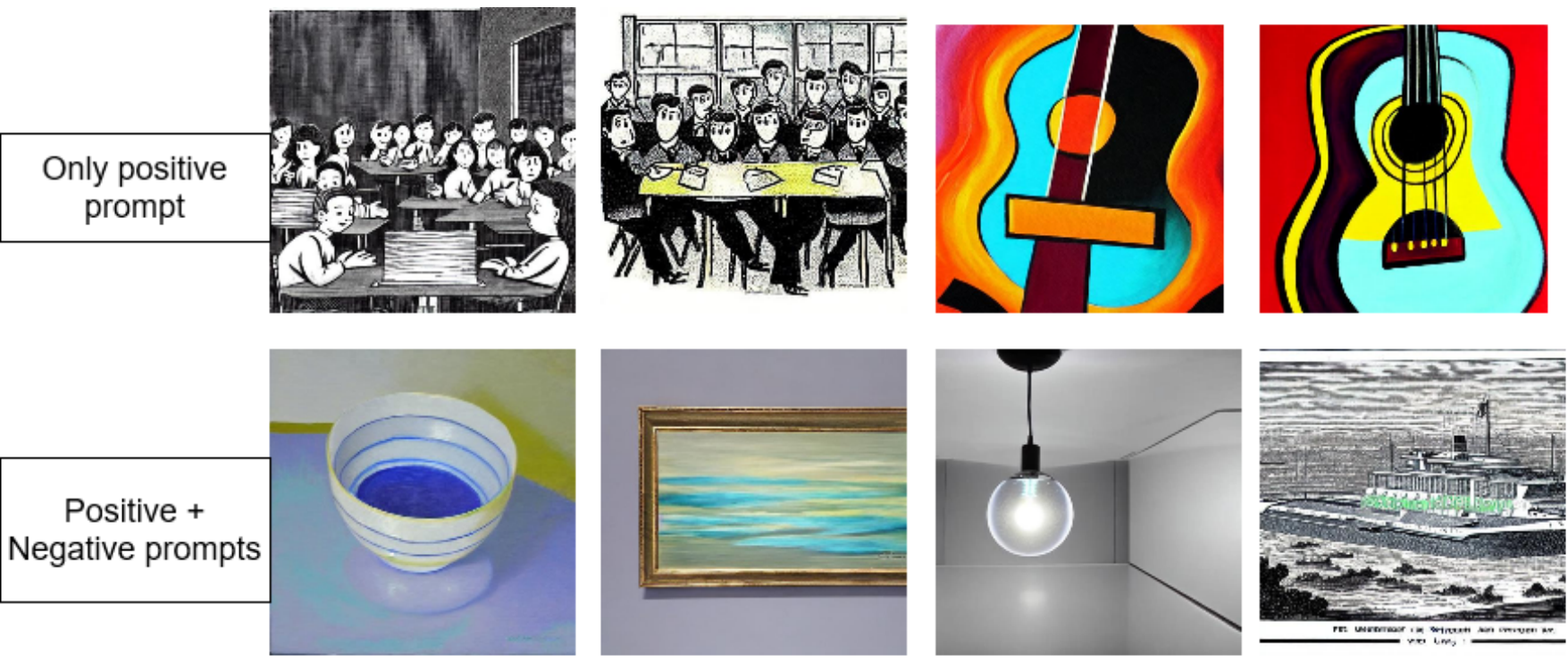}
    \vspace*{-0.5cm}
    \caption{Images generated with only positive prompts vs. both the positive and negative prompts together by stable diffusion.}
    \label{fig:np}
\end{figure}

\noindent \textbf{Effects of \texttt{NP} prompts for pseudo-open image generation}: In Fig. \ref{fig:np}, we note that using only the positive prompt, stable diffusion continues to produce images of known classes. For instance, in the PACS dataset, a positive prompt (\texttt{PP}) repeatedly generates images of 'Person' and 'Guitar', which are inlier classes.

\begin{figure}[htbp!]
    \centering
    \includegraphics[width=\linewidth]{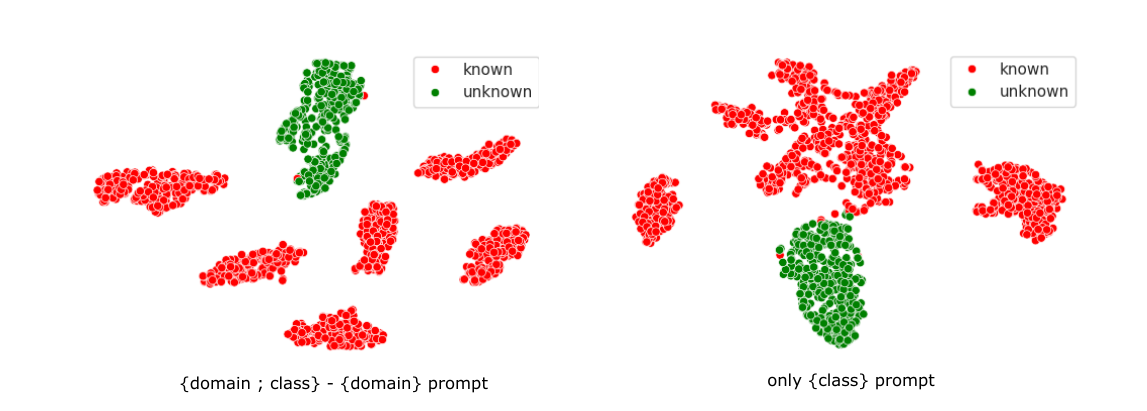}
    \caption{t-SNE of pseudo-open and closed image features produced using a manual $\hat{x}$ and by our proposed approach in ODG-CLIP.}
    \label{fig:tsne1}
\end{figure}

\noindent \textbf{Analysis of the generated visual latent space}:
Figure \ref{fig:tsne1} demonstrates that our method for generating $\tilde{x}$ provides greater discriminability compared to manually defining $\hat{x}$ from static class embeddings.

\begin{figure}[htbp!]
    \centering
    \includegraphics[width=\linewidth]{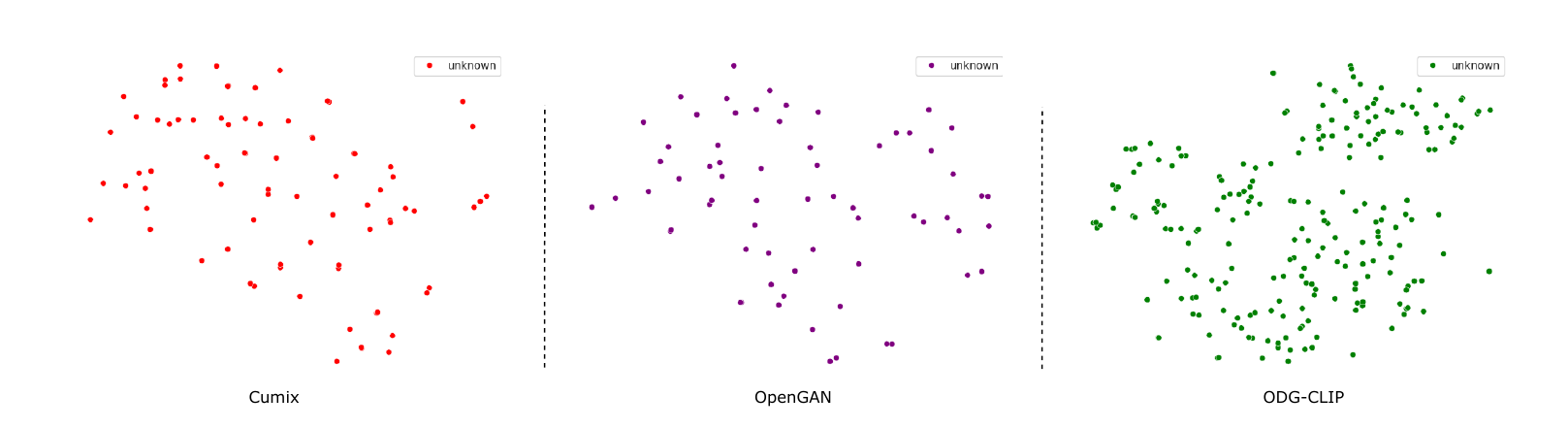}
    \caption{t-SNE of pseudo-open image features produced by Cumix, OpenGAN and ODG-CLIP.}
    \label{fig:tsne2}
    \vspace*{0.2cm}
\end{figure}

\noindent \textbf{t-SNE of open images produced by different methods}: In Figure \ref{fig:tsne2}, we show the t-SNE plots of the CLIP features of the pseudo-open images produced by CuMix, OpenGAN, and the stable diffusion model, which clearly shows that the diffusion based model can better cover the open space.

\section{Model ablation analysis}

Table \ref{tab_encoder} presents the performance outcomes of ODG-CLIP using various CLIP visual encoders.

\begin{table*}[htbp!]
\caption{Ablation with ResNet-50 and ViT/B-16 based CLIP encoders.}
\vspace*{-3mm}
\begin{center}
\begin{adjustbox}{width=\linewidth}
\begin{tabular}{lcccccccccccc|cc}\toprule

{\textbf{Methods}}&\multicolumn{2}{c}{\textbf{PACS}} &\multicolumn{2}{c}{\textbf{VLCS}}&\multicolumn{2}{c}{\textbf{OfficeHome}} &\multicolumn{2}{c}{\textbf{Digits-DG}}&\multicolumn{2}{c}{\textbf{Multi-Dataset}}&\multicolumn{2}{c}{\textbf{Mini DomainNet}}&\multicolumn{2}{|c}{\textbf{Average}}\\
\cmidrule(lr){2-3}\cmidrule(lr){4-5}\cmidrule(lr){6-7}\cmidrule(lr){8-9}\cmidrule(lr){10-11}\cmidrule(lr){12-13}\cmidrule(lr){14-15}
 &\textbf{Acc}&\textbf{H-score}&
 \textbf{Acc}&\textbf{H-score}&
 \textbf{Acc}&\textbf{H-score}&
 \textbf{Acc}&\textbf{H-score}&
 \textbf{Acc}&\textbf{H-score}&
 \textbf{Acc}&\textbf{H-score}&\textbf{Acc}&\textbf{H-score}\\
\midrule

RN-50 &94.30 &90.76 &88.21 &79.36 &91.13 &88.28 &89.76 &72.50 &75.15 &70.27 &88.70 &84.34 &87.88 &80.92 \\

ViT-B/16 &98.64 &97.23 &94.95 &86.24 &97.85 &95.73 &91.44 &77.85 &82.38 &87.62 &94.50 &94.11 &93.29 &89.80 \\

\midrule

\cellcolor{blue!20}\textbf{\textsc{ViT-B/32}} &\cellcolor{blue!20}\cellcolor{blue!20}\textbf{99.53} &\cellcolor{blue!20}\textbf{99.70} &\cellcolor{blue!20}\textbf{95.71}&\cellcolor{blue!20}\textbf{86.53} &\cellcolor{blue!20}\textbf{98.32} &\cellcolor{blue!20}\textbf{96.08}&\cellcolor{blue!20}\textbf{91.53}&\cellcolor{blue!20}\textbf{78.27}&\cellcolor{blue!20}\textbf{84.60}&\cellcolor{blue!20}\textbf{90.00}&\cellcolor{blue!20}\textbf{95.68}&\cellcolor{blue!20}\textbf{94.48}&\cellcolor{blue!20}\textbf{94.23}&\cellcolor{blue!20}\textbf{90.84}\\

\bottomrule
\end{tabular}
\end{adjustbox}
\label{tab_encoder}
\end{center}
\end{table*}

\section{Additional results of  using \texttt{unknown}-class prompts into existing models}

In Table \ref{tab_abl1}, we show further comparisons to the existing prompting techniques, equipped with the \texttt{unknown}-class prompts for the open samples, where the stable-diffusion model \cite{stablediffusion} was used to generate the training pseudo-open images for this prompt. 

\begin{table*}[htbp!]
\caption{Extended comparisons with respect to the prompting techniques coupled with the \texttt{unknown}-class prompts.}
\vspace*{-3mm}
\begin{center}
\begin{adjustbox}{width=\linewidth}
\begin{tabular}{lcccccccccccc|cc}\toprule

{\textbf{Methods}}&\multicolumn{2}{c}{\textbf{PACS}} &\multicolumn{2}{c}{\textbf{VLCS}}&\multicolumn{2}{c}{\textbf{OfficeHome}} &\multicolumn{2}{c}{\textbf{Digits-DG}}&\multicolumn{2}{c}{\textbf{Multi-Dataset}}&\multicolumn{2}{c}{\textbf{Mini DomainNet}}&\multicolumn{2}{|c}{\textbf{Average}}\\
\cmidrule(lr){2-3}\cmidrule(lr){4-5}\cmidrule(lr){6-7}\cmidrule(lr){8-9}\cmidrule(lr){10-11}\cmidrule(lr){12-13}\cmidrule(lr){14-15}
 &\textbf{Acc}&\textbf{H-score}&
 \textbf{Acc}&\textbf{H-score}&
 \textbf{Acc}&\textbf{H-score}&
 \textbf{Acc}&\textbf{H-score}&
 \textbf{Acc}&\textbf{H-score}&
 \textbf{Acc}&\textbf{H-score}&\textbf{Acc}&\textbf{H-score}\\
\midrule

CoOp \cite{coop} + SD \cite{stablediffusion} &92.53 &79.27 &92.24 &70.52 &84.63 &75.34 &80.36 &62.78 &78.10 &71.48 &83.25 &78.55 &85.19 &72.99\\

CoCoOp \cite{cocoop} + SD \cite{stablediffusion} &92.65 &81.45 &92.51 &72.00 &82.35 &79.53 &80.58 &62.95 &78.24 &73.29 &83.50 &78.93 &84.97 &74.69\\ 

MaPLe \cite{maple} + SD \cite{stablediffusion} &91.47 &82.60 &91.70 &72.67 &85.02 &80.60 &79.92 &65.82 &77.62 &72.83 &83.79 &79.30 &84.92 &75.64\\

LASP \cite{lasp} + SD \cite{stablediffusion} &90.32 &82.44 &90.37 &71.19 &81.56 &80.42 &80.55 &62.50 &75.89 &70.04 &82.82 &79.46 &83.59 &74.34\\

PromptSRC \cite{promptsrc} + SD \cite{stablediffusion} &\cellcolor{red!20}93.21 &\cellcolor{red!20}87.95 &90.34 &72.62 &84.60 &\cellcolor{red!20}83.31 &80.92 &65.37 &78.44 &77.89 &83.87 &82.95 &85.23 &78.35\\

\textsc{StyLIP} \cite{stylip} + SD \cite{stablediffusion} &91.78 &87.42 &\cellcolor{red!20}92.11 &\cellcolor{red!20}73.34 &\cellcolor{red!20}85.51 &81.22 &\cellcolor{red!20}81.45 &\cellcolor{red!20}68.10 &\cellcolor{red!20}79.05 &\cellcolor{red!20}78.52 &\cellcolor{red!20}84.12 &\cellcolor{red!20}83.21 &\cellcolor{red!20}85.67&\cellcolor{red!20}78.64\\

\midrule

\cellcolor{blue!20}\textbf{\textsc{ODG-CLIP}} &\cellcolor{blue!20}\cellcolor{blue!20}\textbf{99.53} &\cellcolor{blue!20}\textbf{99.70} &\cellcolor{blue!20}\textbf{95.71}&\cellcolor{blue!20}\textbf{86.53} &\cellcolor{blue!20}\textbf{98.32} &\cellcolor{blue!20}\textbf{96.08}&\cellcolor{blue!20}\textbf{91.53}&\cellcolor{blue!20}\textbf{78.27}&\cellcolor{blue!20}\textbf{84.60}&\cellcolor{blue!20}\textbf{90.00}&\cellcolor{blue!20}\textbf{95.68}&\cellcolor{blue!20}\textbf{94.48}&\cellcolor{blue!20}\textbf{94.23$\pm$0.19}&\cellcolor{blue!20}\textbf{90.84$\pm$0.26}\\

\bottomrule
\end{tabular}
\end{adjustbox}
\label{tab_abl1}
\end{center}
\end{table*}

\section{ODG results on full DomainNet}

In Table \ref{tab_DN}, we show the ODG results on the full DomainNet dataset for all the domain combinations. The dataset splits are mentioned in Table \ref{split}.

\begin{table*}[htbp]
\caption{Comparative analysis for DomainNet in ODG setting on average Acc and H-score over all the domain combinations following leave-one-domain-out protocol.}
\vspace*{-3mm}
\begin{center}
\scalebox{0.6607}{
\begin{tabular}{llcccccccccccc|cc}
\toprule

&\multicolumn{1}{l}{\multirow{2}{*}{\textbf{Methods}}}&\multicolumn{2}{c}{\textbf{Clipart}} &\multicolumn{2}{c}{\textbf{Painting}}&\multicolumn{2}{c}{\textbf{Real}} &\multicolumn{2}{c}{\textbf{Infograph}}&\multicolumn{2}{c}{\textbf{Quickdraw}}&\multicolumn{2}{c}{\textbf{Sketch}}&\multicolumn{2}{|c}{\textbf{Average}}\\
\cmidrule(lr){3-4}\cmidrule(lr){5-6}\cmidrule(lr){7-8}\cmidrule(lr){9-10}\cmidrule(lr){11-12}\cmidrule(lr){13-14}\cmidrule(lr){15-16}
 &&\textbf{Acc}&\textbf{H-score}&
 \textbf{Acc}&\textbf{H-score}&
 \textbf{Acc}&\textbf{H-score}&
 \textbf{Acc}&\textbf{H-score}&
 \textbf{Acc}&\textbf{H-score}&
 \textbf{Acc}&\textbf{H-score}&\textbf{Acc}&\textbf{H-score}\\
\midrule
\multirow{4}{*}{\rotatebox{90}{CNN-based}} &Cumix \cite{cumix} &40.29 &36.23 &29.45 &27.72 &55.76 &44.63 &18.67 &19.53 &6.78 &5.96 &27.43 &28.38 &29.73 &27.08 \\
&MixStyle \cite{mixstyle} &44.24 &38.85 &33.81 &29.68 &58.29 &46.47 &24.18 &21.31 &8.34 &8.62 &34.56 &32.50 &33.90 &29.57\\
&DAML \cite{odg1} &48.59 &46.31 &38.40 &35.25 &59.47 &54.49 &25.63 &25.17 &10.57 &13.00 &35.77 &35.15 &36.41 &34.90\\

&MEDIC \cite{medic} &54.32 &49.33 &40.22 &35.73 &64.60 &53.33 &27.32 &25.27 &9.25 &10.95 &38.12 &37.16 &38.97 &35.29\\
\midrule

\multirow{15}{*}{\rotatebox{90}{CLIP-based}}&\cellcolor{grey}CLIP \cite{clip} &67.98 &52.20 &61.76 &54.00 &82.92 &61.54 &43.50 &43.47 &13.87 &14.46 &55.58 &49.05 &54.27 &45.79 \\

&\cellcolor{grey}CLIP + OpenMax \cite{osr1} &68.05 &36.43 &60.02 &39.85 &80.28 &52.94 &42.41&38.81 &12.48 &13.87 &52.43 &47.97 &52.61 &38.31 \\

&\cellcolor{grey}CLIP + OSDA \cite{osda} &67.45 &37.12 &62.84 &39.02 &82.34 &55.04 &43.84 &39.65 &12.07 &13.66 &53.95 &47.72 &53.75 &38.70 \\

&\cellcolor{cyan!10}CoOp \cite{coop} &68.77 &31.42 &58.94 &26.17 &72.58 &34.11 &45.26 &29.89 &14.71 &10.55 &56.81 &30.93 &52.85 &27.18 \\

&\cellcolor{cyan!10}CoCoOp \cite{cocoop} &66.65 &32.14 &59.94 &20.15 &77.32 &37.01 &46.33 &32.87 &16.82 &13.05 &60.90 &32.53 &54.66 &27.96 \\

&\cellcolor{cyan!10}MaPLe \cite{maple} &74.56 &38.47 &\cellcolor{red!20}67.06 &30.38 &78.14 &42.21 &56.33 &33.55 &12.94 &13.16 &65.97 &38.45 &59.17 &32.70 \\

&\cellcolor{cyan!10}LASP \cite{lasp} &68.20 &36.19 &61.38 &34.08 &75.29 &43.08 &49.81 &34.41 &15.37 &15.13 &62.36 &37.05 &55.40 &33.32 \\

&\cellcolor{cyan!10}PromptSRC \cite{promptsrc} &76.43 &42.55 &66.25 &32.33 &79.17 &43.98 &\cellcolor{red!20}58.29 &36.56 &15.78 &13.93 &66.45 &40.83 &60.40 &35.03 \\

&\cellcolor{cyan!10}CLIPN \cite{clipn} &75.51 &53.40 &62.64 &41.21 &82.49 &56.08 &55.28 &45.37 &17.54 &15.89 &64.58 &48.30 &59.67 &43.37 \\

&\cellcolor{cyan!10}\textsc{StyLIP} \cite{stylip} &79.14 &48.23 &64.80 &46.39 &\cellcolor{red!20}86.52 &53.07 &56.12 &42.74 &18.65 &16.85 &68.14 &45.48 &62.23 &42.13 \\

&\cellcolor{cyan!10}CLIPN + \textsc{StyLIP} &78.67 &57.41 &65.22 &46.73 &84.20 &57.20 &53.48 &38.22 &\cellcolor{red!20}18.78 &\cellcolor{red!20}17.75 &67.93 &49.95 &61.38 &44.54 \\

&\cellcolor{violet!20}MaPLe + SD &75.22 &66.86 &64.21 &56.40 &79.27 &69.10 &55.25 &53.77 &13.46 &13.95 &66.15 &58.95 &58.93 &53.17\\

&\cellcolor{violet!20}PromptSRC + SD &75.39 &68.92 &62.48 &\cellcolor{red!20}60.34 &79.93 &70.76 &57.82 &\cellcolor{red!20}57.01 &16.38 &15.89 &\cellcolor{red!20}69.37 &61.50 &60.23 &55.74\\

&\cellcolor{violet!20}\textsc{StyLIP} + SD &\cellcolor{red!20}79.25 &\cellcolor{red!20}71.60 &65.04 &59.14 &85.19 &\cellcolor{red!20}74.45 &56.73 &55.16 &17.32 &17.18 &68.93 &\cellcolor{red!20}62.60 &\cellcolor{red!20}62.08 &\cellcolor{red!20}56.69\\

\cmidrule(lr){2-16}

&\cellcolor{blue!20}\textbf{\textsc{ODG-CLIP}} &\cellcolor{blue!20}\cellcolor{blue!20}\textbf{90.41} &\cellcolor{blue!20}\textbf{85.07} &\cellcolor{blue!20}\textbf{79.28}&\cellcolor{blue!20}\textbf{75.19} &\cellcolor{blue!20}\textbf{92.38} &\cellcolor{blue!20}\textbf{87.63}&\cellcolor{blue!20}\textbf{65.34}&\cellcolor{blue!20}\textbf{66.80}&\cellcolor{blue!20}\textbf{25.41}&\cellcolor{blue!20}\textbf{25.47}&\cellcolor{blue!20}\textbf{78.46}&\cellcolor{blue!20}\textbf{73.65}&\cellcolor{blue!20}\textbf{71.88}&\cellcolor{blue!20}\textbf{68.97}\\
\bottomrule
\end{tabular}}
\label{tab_DN}
\end{center}
\end{table*}

\section{Complete results on the all the datasets for ODG and closed-set DG}

Please refer to Tables \ref{tab_pacs_open}-\ref{tab_minidomainnet_open} for the detailed ODG results and Table \ref{tab_pacsvlcsoh_closed}-\ref{tab_digitsmini_closed} for the closed-set DG results, respectively.

\begin{table*}[htbp]
\caption{Comparative analysis for PACS in ODG setting on average Acc and H-score over all the domain combinations following leave-one-domain-out protocol.}
\vspace*{-3mm}
\begin{center}
\scalebox{0.6607}{
\begin{tabular}{llcccccccc|cc}
\toprule

&\multicolumn{1}{l}{\multirow{2}{*}{\textbf{Methods}}}&\multicolumn{2}{c}{\textbf{Art}} &\multicolumn{2}{c}{\textbf{Sketch}}&\multicolumn{2}{c}{\textbf{Photo}} &\multicolumn{2}{c}{\textbf{Cartoon}}&\multicolumn{2}{|c}{\textbf{Average}}\\
\cmidrule(lr){3-4}\cmidrule(lr){5-6}\cmidrule(lr){7-8}\cmidrule(lr){9-10}\cmidrule(lr){11-12}
 &&\textbf{Acc}&\textbf{H-score}&
 \textbf{Acc}&\textbf{H-score}&
 \textbf{Acc}&\textbf{H-score}&
 \textbf{Acc}&\textbf{H-score}&\textbf{Acc}&\textbf{H-score}\\
\midrule
\multirow{4}{*}{\rotatebox{90}{CNN-based}} &Cumix \cite{cumix} & 53.85 & 38.67 & 37.70 & 28.71 & 65.67 & 49.28 & 74.16 & 47.53 & 57.85 & 41.05 \\
&MixStyle \cite{mixstyle} &53.41 & 39.33 & 56.10 & 54.44 & 72.37 & 47.21 & 71.54 & 52.22 & 63.36 & 48.30 \\
&DAML \cite{odg1} &54.10 & 43.02 & 58.50 & 56.73 & 75.69 & 53.29 & 73.65 & 54.47 & 65.49 & 51.88 \\

&MEDIC \cite{medic} & 91.62 & 81.61 & 84.61 & 78.35 & 96.37 & 94.75 & 86.65 & 77.39 & 89.81 & 83.03 \\
\midrule

\multirow{15}{*}{\rotatebox{90}{CLIP-based}}&\cellcolor{grey}CLIP \cite{clip} & 96.87 & 73.50 & 85.38 & 70.90 & 99.75 & 92.83 & 98.65 & 69.85 & 95.16 & 76.77 \\

&\cellcolor{grey}CLIP + OpenMax \cite{osr1} & 95.25 & 76.19 & 85.27 & 72.15 & 96.18 & 95.60 & 97.10 & 72.59 & 93.45 & 79.13 \\

&\cellcolor{grey}CLIP + OSDA \cite{osda} & 93.48 & 73.38 & 85.46 & 67.64 & 95.26 & 92.29 & 96.28 & 68.30 & 92.62 & 75.40 \\

&\cellcolor{cyan!10}CoOp \cite{coop} & 96.23 & 29.60 & 83.05 & 21.91 & 89.04 & 34.78 & 46.77 & 21.20 & 78.77 & 26.87 \\

&\cellcolor{cyan!10}CoCoOp \cite{cocoop} & 95.17 & 30.81 & 84.77 & 22.54 & 90.30 & 40.15 & 72.80 & 38.23 & 85.76 & 32.93 \\

&\cellcolor{cyan!10}MaPLe \cite{maple} & 95.70 & 37.89 & 85.69 & 26.42 & 99.03 & 68.46 & 95.46 & 61.12 & 93.97 & 48.47 \\

&\cellcolor{cyan!10}LASP \cite{lasp} & 95.34 & 28.45 & 86.38 & 22.56 & 93.48 & 36.29 & 78.61 & 34.19 & 88.45 & 30.37 \\

&\cellcolor{cyan!10}PromptSRC \cite{promptsrc} & 96.05 & 30.14 & 87.23 & 23.49 & 98.6 & 62.36 & 96.24 & 57.27 & 94.53 & 43.32 \\

&\cellcolor{cyan!10}CLIPN \cite{clipn} & 97.27 & 32.50 & 91.71 & 20.80 & 98.15 & 66.17 & 97.83 & 60.52 & 96.24 & 45.00 \\

&\cellcolor{cyan!10}\textsc{StyLIP} \cite{stylip} & 96.93 & 40.74 & 92.34 & 28.51 & 96.38 & 70.43 & 95.79 & 63.26 & 95.36 & 50.74 \\

&\cellcolor{cyan!10}CLIPN + \textsc{StyLIP} & 97.05 & 59.27 & 91.86 & 42.78 & 98.44 & 77.65 & 98.13 & 78.13 & 96.37 & 64.46 \\

&\cellcolor{violet!20}MaPLe + SD &94.35	&84.79	&84.42	&74.13	&95.25	&85.76	&91.87	&85.70	&91.47	&82.60 \\

&\cellcolor{violet!20}PromptSRC + SD & 94.84 & 88.51 & 89.30 & 83.59 & 94.28 & 90.35 & 94.43 & 89.36 & 93.21 & 87.95\\

&\cellcolor{violet!20}\textsc{StyLIP} + SD & 95.27 & 87.48 & 87.25 & 81.38 & 91.65 & 90.93 & 92.95 & 89.90 & 91.78 & 87.42 \\

\cmidrule(lr){2-12}

&\cellcolor{blue!20}\textbf{\textsc{ODG-CLIP}} &\cellcolor{blue!20}\cellcolor{blue!20}\textbf{99.42} &\cellcolor{blue!20}\textbf{99.58} &\cellcolor{blue!20}\textbf{99.17}&\cellcolor{blue!20}\textbf{99.67} &\cellcolor{blue!20}\textbf{100.00} &\cellcolor{blue!20}\textbf{100.00}&\cellcolor{blue!20}\textbf{99.52}&\cellcolor{blue!20}\textbf{99.54}&\cellcolor{blue!20}\textbf{99.53}&\cellcolor{blue!20}\textbf{99.70}\\
\bottomrule
\end{tabular}}
\label{tab_pacs_open}
\end{center}
\end{table*}

\begin{table*}[htbp]
\caption{Comparative analysis for VLCS in ODG setting on average Acc and H-score over all the domain combinations following leave-one-domain-out protocol.}
\vspace*{-3mm}
\begin{center}
\scalebox{0.6607}{
\begin{tabular}{llcccccccc|cc}
\toprule
&\multicolumn{1}{l}{\multirow{2}{*}{\textbf{Methods}}}&\multicolumn{2}{c}{\textbf{Caltech}} &\multicolumn{2}{c}{\textbf{LabelMe}}&\multicolumn{2}{c}{\textbf{Pascal VOC}} &\multicolumn{2}{c}{\textbf{Sun}}&\multicolumn{2}{|c}{\textbf{Average}}\\
\cmidrule(lr){3-4}\cmidrule(lr){5-6}\cmidrule(lr){7-8}\cmidrule(lr){9-10}\cmidrule(lr){11-12}
 &&\textbf{Acc}&\textbf{H-score}&
 \textbf{Acc}&\textbf{H-score}&
 \textbf{Acc}&\textbf{H-score}&
 \textbf{Acc}&\textbf{H-score}&\textbf{Acc}&\textbf{H-score}\\
\midrule
\multirow{4}{*}{\rotatebox{90}{CNN-based}} &Cumix \cite{cumix} & 66.21 & 63.76 & 46.72 & 45.59 & 50.54 & 45.78 & 46.38 & 45.32 & 52.46 & 50.11 \\
&MixStyle \cite{mixstyle} & 66.11 & 63.19 & 46.72 & 46.22 & 49.75 & 46.19 & 46.62 & 46.85 & 52.30 & 50.61 \\
&DAML \cite{odg1} & 69.18 & 64.65 & 48.22 & 47.71 & 49.87 & 47.22 & 46.87 & 46.78 & 53.54 & 51.59 \\
&MEDIC \cite{medic} & 76.47 & 69.90 & 52.47 & 55.27 & 52.91 & 50.43 & 47.25 & 47.32 & 57.28 & 55.73 \\
\midrule

\multirow{15}{*}{\rotatebox{90}{CLIP-based}}&\cellcolor{grey}CLIP \cite{clip} & 97.32 & 83.33 & 92.54 & 73.03 & 86.28 & 62.93 & 91.20 & 72.48 & 91.84 & 72.94 \\

&\cellcolor{grey}CLIP + OpenMax \cite{osr1} & 97.92 & 85.25 & 93.67 & 76.51 & 85.98 & 62.34 & 90.78 & 70.57 & 92.09 & 73.67 \\

&\cellcolor{grey}CLIP + OSDA \cite{osda} & 96.53 & 80.36 & 90.23 & 72.43 & 82.45 & 60.55 & 91.64 & 70.23 & 90.21 & 70.89 \\

&\cellcolor{cyan!10}CoOp \cite{coop} & 98.17 & 38.00 & 91.74 & 36.64 & 87.37 & 34.79 & 90.79 & 47.60 & 92.02 & 39.26 \\

&\cellcolor{cyan!10}CoCoOp \cite{cocoop} & 96.86 & 30.70 & 87.11 & 37.78 & 87.52 & 34.30 & 86.40 & 45.27 & 89.47 & 37.01 \\

&\cellcolor{cyan!10}MaPLe \cite{maple} & 93.72 & 45.92 & 90.53 & 43.18 & 86.07 & 48.50 & 88.46 & 35.71 & 89.70 & 43.33 \\

&\cellcolor{cyan!10}LASP \cite{lasp} & 95.37 & 39.54 & 88.62 & 39.47 & 89.40 & 47.12 & 89.28 & 31.51 & 90.67 & 39.41 \\

&\cellcolor{cyan!10}PromptSRC \cite{promptsrc} & 94.92 & 40.47 & 91.37 & 44.27 & 86.66 & 51.37 & 87.55 & 35.00 & 90.13 & 42.78 \\

&\cellcolor{cyan!10}CLIPN \cite{clipn} & 92.47 & 59.36 & 84.19 & 50.59 & 80.48 & 59.20 & 82.13 & 33.73 & 84.82 & 50.72 \\

&\cellcolor{cyan!10}\textsc{StyLIP} \cite{stylip} & 96.26 & 70.35 & 92.48 & 68.25 & 87.22 & 65.32 & 87.05 & 58.71 & 90.75 & 65.66 \\

&\cellcolor{cyan!10}CLIPN + \textsc{StyLIP} & 92.31 & 73.68 & 85.50 & 71.46 & 80.42 & 68.79 & 80.35 & 58.15 & 84.65 & 68.02 \\

&\cellcolor{violet!20}MaPLe + SD & 96.45 & 79.26 & 93.24 & 75.24 & 88.82 & 70.25 & 88.30 & 65.94 & 91.70 & 72.67 \\

&\cellcolor{violet!20}PromptSRC + SD & 96.02 & 79.94 & 90.66 & 73.10 & 88.20 & 70.94 & 86.47 & 66.50 & 90.34 & 72.62\\

&\cellcolor{violet!20}\textsc{StyLIP} + SD & 97.64 & 80.82 & 94.25 & 75.95 & 90.23 & 70.20 & 86.30 & 66.37 & 92.11 & 73.34 \\

\cmidrule(lr){2-12}

&\cellcolor{blue!20}\textbf{\textsc{ODG-CLIP}} &\cellcolor{blue!20}\cellcolor{blue!20}\textbf{98.35} &\cellcolor{blue!20}\textbf{90.75} &\cellcolor{blue!20}\textbf{96.35}&\cellcolor{blue!20}\textbf{89.45} &\cellcolor{blue!20}\textbf{94.65} &\cellcolor{blue!20}\textbf{88.05}&\cellcolor{blue!20}\textbf{93.48}&\cellcolor{blue!20}\textbf{77.85}&\cellcolor{blue!20}\textbf{95.71}&\cellcolor{blue!20}\textbf{86.53}\\
\bottomrule
\end{tabular}}
\label{tab_vlcs_open}
\end{center}
\end{table*}

\begin{table*}[htbp]
\caption{Comparative analysis for Office-Home in ODG setting on average Acc and H-score over all the domain combinations following leave-one-domain-out protocol.}
\vspace*{-3mm}
\begin{center}
\scalebox{0.6607}{
\begin{tabular}{llcccccccc|cc}
\toprule

&\multicolumn{1}{l}{\multirow{2}{*}{\textbf{Methods}}}&\multicolumn{2}{c}{\textbf{Clipart}} &\multicolumn{2}{c}{\textbf{Real-World}}&\multicolumn{2}{c}{\textbf{Product}} &\multicolumn{2}{c}{\textbf{Art}}&\multicolumn{2}{|c}{\textbf{Average}}\\
\cmidrule(lr){3-4}\cmidrule(lr){5-6}\cmidrule(lr){7-8}\cmidrule(lr){9-10}\cmidrule(lr){11-12}
 &&\textbf{Acc}&\textbf{H-score}&
 \textbf{Acc}&\textbf{H-score}&
 \textbf{Acc}&\textbf{H-score}&
 \textbf{Acc}&\textbf{H-score}&\textbf{Acc}&\textbf{H-score}\\
\midrule
\multirow{4}{*}{\rotatebox{90}{CNN-based}} &Cumix \cite{cumix} & 41.54 & 43.07 & 64.63 & 58.02 & 57.74 & 55.79 & 42.76 & 40.72 & 51.67 & 49.40 \\
&MixStyle \cite{mixstyle} & 42.28 & 41.15 & 61.78 & 60.23 & 59.92 & 53.97 & 50.11 & 42.78 & 53.52 & 49.53 \\
&DAML \cite{odg1} & 45.13 & 43.12 & 65.99 & 60.13 & 61.54 & 59.00 & 53.13 & 51.11 & 56.45 & 53.34 \\
&MEDIC \cite{medic} & 48.96 & 49.39 & 67.42 & 61.00 & 65.20 & 66.09 & 59.46 & 55.17 & 60.26 & 57.91 \\
\midrule

\multirow{15}{*}{\rotatebox{90}{CLIP-based}}&\cellcolor{grey}CLIP \cite{clip} & 68.07 & 64.02 & 90.02 & 67.35 & 86.79 & 57.77 & 80.82 & 65.34 & 81.43 & 63.62 \\

&\cellcolor{grey}CLIP + OpenMax \cite{osr1} & 68.44 & 63.41 & 89.10 & 62.30 & 85.25 & 55.32 & 81.20 & 65.12 & 81.00 & 61.54 \\

&\cellcolor{grey}CLIP + OSDA \cite{osda} & 69.76 & 67.93 & 91.67 & 70.65 & 84.60 & 61.53 & 84.29 & 69.28 & 82.58 & 67.35 \\

&\cellcolor{cyan!10}CoOp \cite{coop} & 65.28 & 39.54 & 82.07 & 36.04 & 79.02 & 30.91 & 69.03 & 38.55 & 73.85 & 36.26 \\

&\cellcolor{cyan!10}CoCoOp \cite{cocoop} & 68.21 & 33.05 & 81.62 & 39.41 & 80.92 & 30.19 & 70.77 & 34.86 & 75.38 & 34.38 \\

&\cellcolor{cyan!10}MaPLe \cite{maple} & 79.48 & 36.57 & 85.44 & 31.42 & 77.11 & 28.23 & 75.83 & 36.00 & 79.47 & 33.06 \\

&\cellcolor{cyan!10}LASP \cite{lasp} & 72.36 & 32.75 & 80.50 & 37.78 & 76.37 & 31.38 & 75.27 & 36.15 & 76.13 & 34.52 \\

&\cellcolor{cyan!10}PromptSRC \cite{promptsrc} & 80.27 & 38.26 & 86.25 & 36.27 & 78.30 & 32.47 & 76.01 & 38.58 & 80.21 & 36.40 \\

&\cellcolor{cyan!10}CLIPN \cite{clipn} & 84.18 & 86.54 & 89.47 & 28.53 & 85.45 & 28.20 & 79.10 & 28.05 & 84.55 & 42.83 \\

&\cellcolor{cyan!10}\textsc{StyLIP} \cite{stylip} & 86.32 & 45.56 & 88.35 & 65.38 & 84.92 & 65.62 & 79.33 & 67.32 & 84.73 & 60.97 \\

&\cellcolor{cyan!10}CLIPN + \textsc{StyLIP} & 85.97 & 84.71 & 85.69 & 70.31 & 84.10 & 72.02 & 78.92 & 78.94 & 83.67 & 76.50 \\

&\cellcolor{violet!20}MaPLe + SD & 87.23 & 81.32 & 89.34 & 80.79 & 84.15 & 81.00 & 79.34 & 79.30 & 85.02 & 80.60 \\

&\cellcolor{violet!20}PromptSRC + SD & 87.37 & 83.27 & 89.37 & 84.26 & 85.40 & 82.45 & 80.24 & 83.25 & 85.60 & 83.31\\

&\cellcolor{violet!20}\textsc{StyLIP} + SD & 90.36 & 83.02 & 89.26 & 80.93 & 83.92 & 80.47 & 78.50 & 80.44 & 85.51 & 81.22 \\

\cmidrule(lr){2-12}

&\cellcolor{blue!20}\textbf{\textsc{ODG-CLIP}} &\cellcolor{blue!20}\cellcolor{blue!20}\textbf{97.84} &\cellcolor{blue!20}\textbf{96.33} &\cellcolor{blue!20}\textbf{98.74}&\cellcolor{blue!20}\textbf{95.36} &\cellcolor{blue!20}\textbf{99.50} &\cellcolor{blue!20}\textbf{96.74}&\cellcolor{blue!20}\textbf{97.18}&\cellcolor{blue!20}\textbf{95.88}&\cellcolor{blue!20}\textbf{98.32}&\cellcolor{blue!20}\textbf{96.08}\\
\bottomrule
\end{tabular}}
\label{tab_officehome_open}
\end{center}
\end{table*}

\begin{table*}[htbp]
\caption{Comparative analysis for Digits-DG in ODG setting on average Acc and H-score over all the domain combinations following leave-one-domain-out protocol.}
\vspace*{-3mm}
\begin{center}
\scalebox{0.6607}{
\begin{tabular}{llcccccccc|cc}
\toprule

&\multicolumn{1}{l}{\multirow{2}{*}{\textbf{Methods}}}&\multicolumn{2}{c}{\textbf{MNIST}} &\multicolumn{2}{c}{\textbf{MNIST-M}}&\multicolumn{2}{c}{\textbf{SVHN}} &\multicolumn{2}{c}{\textbf{SYN}}&\multicolumn{2}{|c}{\textbf{Average}}\\
\cmidrule(lr){3-4}\cmidrule(lr){5-6}\cmidrule(lr){7-8}\cmidrule(lr){9-10}\cmidrule(lr){11-12}
 &&\textbf{Acc}&\textbf{H-score}&
 \textbf{Acc}&\textbf{H-score}&
 \textbf{Acc}&\textbf{H-score}&
 \textbf{Acc}&\textbf{H-score}&\textbf{Acc}&\textbf{H-score}\\
\midrule
\multirow{4}{*}{\rotatebox{90}{CNN-based}} &Cumix \cite{cumix} & 72.10 & 67.52 & 45.88 & 43.74 & 52.22 & 47.22 & 62.33 & 58.33 & 58.13 & 54.20 \\
&MixStyle \cite{mixstyle} & 76.56 & 70.56 & 47.81 & 45.66 & 54.97 & 47.24 & 61.80 & 61.96 & 60.29 & 56.36 \\
&DAML \cite{odg1} & 73.98 & 69.88 & 46.49 & 45.62 & 53.34 & 47.72 & 64.22 & 59.23 & 59.51 & 55.61 \\
&MEDIC \cite{medic} & 97.89 & 83.20 & 71.14 & 60.98 & 76.00 & 58.77 & 88.11 & 62.24 & 83.29 & 66.30 \\
\midrule

\multirow{15}{*}{\rotatebox{90}{CLIP-based}}&\cellcolor{grey}CLIP \cite{clip} & 80.35 & 73.73 & 67.83 & 53.82 & 70.83 & 59.62 & 89.31 & 60.63 & 77.08 & 61.95 \\

&\cellcolor{grey}CLIP + OpenMax \cite{osr1} & 79.28 & 76.32 & 63.49 & 51.18 & 74.30 & 60.83 & 90.65 & 62.78 & 76.93 & 62.78 \\

&\cellcolor{grey}CLIP + OSDA \cite{osda} & 81.54 & 79.51 & 71.50 & 54.21 & 78.91 & 64.11 & 90.17 & 64.95 & 80.53 & 65.70 \\

&\cellcolor{cyan!10}CoOp \cite{coop} & 72.98 & 48.06 & 44.29 & 30.09 & 47.02 & 29.67 & 69.88 & 31.43 & 58.54 & 34.81 \\

&\cellcolor{cyan!10}CoCoOp \cite{cocoop} & 45.24 & 41.01 & 50.60 & 28.96 & 49.29 & 31.42 & 65.95 & 32.62 & 52.77 & 33.50 \\

&\cellcolor{cyan!10}MaPLe \cite{maple} & 77.74 & 55.19 & 58.21 & 37.35 & 61.67 & 43.52 & 84.52 & 39.25 & 70.54 & 43.83 \\

&\cellcolor{cyan!10}LASP \cite{lasp} & 61.43 & 42.65 & 51.32 & 29.30 & 51.33 & 38.70 & 79.48 & 30.27 & 60.89 & 35.23 \\

&\cellcolor{cyan!10}PromptSRC \cite{promptsrc} & 85.31 & 57.20 & 63.32 & 40.22 & 63.95 & 43.87 & 88.79 & 35.72 & 75.34 & 44.25 \\

&\cellcolor{cyan!10}CLIPN \cite{clipn} & 93.80 & 58.37 & 70.18 & 42.49 & 72.47 & 45.91 & 90.35 & 35.46 & 81.70 & 45.56 \\

&\cellcolor{cyan!10}\textsc{StyLIP} \cite{stylip} & 94.29 & 70.51 & 70.03 & 50.37 & 68.50 & 61.12 & 89.54 & 50.61 & 80.59 & 58.15 \\

&\cellcolor{cyan!10}CLIPN + \textsc{StyLIP} & 93.87 & 71.43 & 69.74 & 51.28 & 74.52 & 60.84 & 90.43 & 53.42 & 82.14 & 59.24 \\

&\cellcolor{violet!20}MaPLe + SD & 91.44 & 77.19 & 67.92 & 59.59 & 73.33 & 66.28 & 86.97 & 60.21 & 79.92 & 65.82 \\

&\cellcolor{violet!20}PromptSRC + SD & 92.80 & 75.24 & 67.13 & 57.70 & 75.11 & 66.50 & 88.63 & 62.03 & 80.92 & 65.37 \\

&\cellcolor{violet!20}\textsc{StyLIP} + SD & 93.68 & 78.73 & 69.84 & 60.35 & 75.23 & 68.21 & 87.05 & 65.12 & 81.45 & 68.10 \\

\cmidrule(lr){2-12}

&\cellcolor{blue!20}\textbf{\textsc{ODG-CLIP}} &\cellcolor{blue!20}\cellcolor{blue!20}\textbf{96.24} &\cellcolor{blue!20}\textbf{87.14} &\cellcolor{blue!20}\textbf{86.23}&\cellcolor{blue!20}\textbf{72.10} &\cellcolor{blue!20}\textbf{87.41} &\cellcolor{blue!20}\textbf{79.34}&\cellcolor{blue!20}\textbf{96.24}&\cellcolor{blue!20}\textbf{74.51}&\cellcolor{blue!20}\textbf{91.53}&\cellcolor{blue!20}\textbf{78.27}\\
\bottomrule
\end{tabular}}
\label{tab_digitsdg_open}
\end{center}
\end{table*}

\begin{table*}[htbp]
\caption{Comparative analysis for Multi Dataset in ODG setting on average Acc and H-score over all the domain combinations following leave-one-domain-out protocol.}
\vspace*{-3mm}
\begin{center}
\scalebox{0.6607}{
\begin{tabular}{llcccccccc|cc}
\toprule

&\multicolumn{1}{l}{\multirow{2}{*}{\textbf{Methods}}}&\multicolumn{2}{c}{\textbf{Clipart}} &\multicolumn{2}{c}{\textbf{Real}}&\multicolumn{2}{c}{\textbf{Painting}} &\multicolumn{2}{c}{\textbf{Sketch}}&\multicolumn{2}{|c}{\textbf{Average}}\\
\cmidrule(lr){3-4}\cmidrule(lr){5-6}\cmidrule(lr){7-8}\cmidrule(lr){9-10}\cmidrule(lr){11-12}
 &&\textbf{Acc}&\textbf{H-score}&
 \textbf{Acc}&\textbf{H-score}&
 \textbf{Acc}&\textbf{H-score}&
 \textbf{Acc}&\textbf{H-score}&\textbf{Acc}&\textbf{H-score}\\
\midrule
\multirow{4}{*}{\rotatebox{90}{CNN-based}} &Cumix \cite{cumix} & 30.03 & 40.18 & 64.61 & 65.07 & 44.37 & 48.70 & 29.72 & 33.70 & 42.18 & 46.91 \\
&MixStyle \cite{mixstyle} & 31.24 & 38.56 & 65.32 & 66.25 & 44.72 & 47.32 & 27.43 & 35.49 & 42.18 & 46.91 \\
&DAML \cite{odg1} & 37.62 & 44.27 & 66.54 & 67.80 & 47.80 & 52.93 & 34.48 & 41.82 & 46.61 & 51.71 \\
&MEDIC \cite{medic} & 43.13 & 36.74 & 68.87 & 68.14 & 50.93 & 55.21 & 40.02 & 52.41 & 50.74 & 53.13 \\
\midrule

\multirow{15}{*}{\rotatebox{90}{CLIP-based}}&\cellcolor{grey}CLIP \cite{clip} & 81.00 & 74.13 & 84.02 & 72.31 & 69.53 & 68.77 & 76.98 & 73.55 & 77.88 & 72.19 \\

&\cellcolor{grey}CLIP + OpenMax \cite{osr1} & 81.45 & 75.32 & 84.68 & 73.69 & 70.21 & 69.19 & 77.03 & 74.83 & 78.34 & 73.26 \\

&\cellcolor{grey}CLIP + OSDA \cite{osda} & 75.21 & 78.41 & 80.29 & 76.56 & 68.92 & 66.32 & 73.37 & 79.57 & 74.45 & 75.22 \\

&\cellcolor{cyan!10}CoOp \cite{coop} & 66.00 & 51.65 & 63.11 & 38.72 & 69.90 & 45.97 & 65.10 & 41.03 & 66.03 & 44.34 \\

&\cellcolor{cyan!10}CoCoOp \cite{cocoop} & 68.76 & 55.99 & 60.18 & 44.52 & 67.86 & 47.01 & 62.57 & 42.77 & 64.84 & 47.57 \\

&\cellcolor{cyan!10}MaPLe \cite{maple} & 72.42 & 67.51 & 65.49 & 56.00 & 73.20 & 64.35 & 66.25 & 60.93 & 69.34 & 62.20 \\

&\cellcolor{cyan!10}LASP \cite{lasp} & 71.90 & 56.20 & 62.06 & 49.15 & 69.25 & 48.73 & 63.90 & 46.78 & 66.78 & 50.22 \\

&\cellcolor{cyan!10}PromptSRC \cite{promptsrc}& 73.15 & 64.29 & 61.75 & 55.39 & 64.72 & 60.45 & 62.41 & 57.67 & 65.51 & 59.45 \\

&\cellcolor{cyan!10}CLIPN \cite{clipn} & 80.39 & 68.43 & 71.99 & 58.40 & 80.61 & 65.10 & 75.63 & 58.46 & 77.16 & 62.60 \\

&\cellcolor{cyan!10}\textsc{StyLIP} \cite{stylip} & 83.25 & 80.30 & 74.32 & 70.52 & 82.89 & 76.83 & 79.07 & 60.32 & 79.88 & 71.99 \\

&\cellcolor{cyan!10}CLIPN + \textsc{StyLIP} & 82.36 & 81.57 & 70.79 & 68.84 & 80.47 & 77.50 & 74.10 & 60.67 & 76.93 & 72.15 \\

&\cellcolor{violet!20}MaPLe + SD & 82.93 & 82.43 & 71.55 & 69.31 & 81.59 & 78.21 & 74.39 & 61.36 & 77.62 & 72.83 \\

&\cellcolor{violet!20}PromptSRC + SD & 83.54 & 87.31 & 72.40 & 76.09 & 84.13 & 84.58 & 73.68 & 63.59 & 78.44 & 77.89 \\

&\cellcolor{violet!20}\textsc{StyLIP} + SD & 84.30 & 87.78 & 73.75 & 76.56 & 85.92 & 86.38 & 72.21 & 63.34 & 79.05 & 78.52 \\

\cmidrule(lr){2-12}

&\cellcolor{blue!20}\textbf{\textsc{ODG-CLIP}} &\cellcolor{blue!20}\cellcolor{blue!20}\textbf{90.65} &\cellcolor{blue!20}\textbf{94.43} &\cellcolor{blue!20}\textbf{80.39}&\cellcolor{blue!20}\textbf{88.31} &\cellcolor{blue!20}\textbf{90.47} &\cellcolor{blue!20}\textbf{91.53}&\cellcolor{blue!20}\textbf{76.89}&\cellcolor{blue!20}\textbf{85.72}&\cellcolor{blue!20}\textbf{84.60}&\cellcolor{blue!20}\textbf{90.00}\\
\bottomrule
\end{tabular}}
\label{tab_multidata_open}
\end{center}
\end{table*}

\begin{table*}[htbp]
\caption{Comparative analysis for Mini-DomainNet in ODG setting on average Acc and H-score over all the domain combinations following leave-one-domain-out protocol.}
\vspace*{-3mm}
\begin{center}
\scalebox{0.6607}{
\begin{tabular}{llcccccccc|cc}
\toprule

&\multicolumn{1}{l}{\multirow{2}{*}{\textbf{Methods}}}&\multicolumn{2}{c}{\textbf{Clipart}} &\multicolumn{2}{c}{\textbf{Real}}&\multicolumn{2}{c}{\textbf{Painting}} &\multicolumn{2}{c}{\textbf{Sketch}}&\multicolumn{2}{|c}{\textbf{Average}}\\
\cmidrule(lr){3-4}\cmidrule(lr){5-6}\cmidrule(lr){7-8}\cmidrule(lr){9-10}\cmidrule(lr){11-12}
 &&\textbf{Acc}&\textbf{H-score}&
 \textbf{Acc}&\textbf{H-score}&
 \textbf{Acc}&\textbf{H-score}&
 \textbf{Acc}&\textbf{H-score}&\textbf{Acc}&\textbf{H-score}\\
\midrule
\multirow{4}{*}{\rotatebox{90}{CNN-based}} &Cumix \cite{cumix} & 46.48 & 30.50 & 62.13 & 53.58 & 54.02 & 47.54 & 38.46 & 25.00 & 50.27 & 39.16 \\
&MixStyle \cite{mixstyle} & 46.59 & 31.39 & 63.56 & 55.69 & 55.15 & 48.45 & 36.42 & 25.45 & 50.43 & 40.25 \\
&DAML \cite{odg1} & 47.39 & 36.21 & 67.37 & 58.21 & 60.37 & 50.58 & 36.11 & 29.52 & 52.81 & 43.63 \\
&MEDIC \cite{medic} & 51.98 & 38.36 & 67.53 & 60.12 & 65.32 & 51.78 & 36.32 & 32.56 & 55.29 & 45.71 \\
\midrule

\multirow{15}{*}{\rotatebox{90}{CLIP-based}}&\cellcolor{grey}CLIP \cite{clip} & 88.00 & 69.35 & 90.50 & 68.84 & 80.00 & 66.72 & 79.50 & 70.85 & 84.50 & 68.94 \\

&\cellcolor{grey}CLIP + OpenMax \cite{osr1} & 85.36 & 71.47 & 89.44 & 67.47 & 77.20 & 68.21 & 75.56 & 70.46 & 81.89 & 69.40 \\

&\cellcolor{grey}CLIP + OSDA \cite{osda} & 86.32 & 76.32 & 88.57 & 70.31 & 81.34 & 74.59 & 71.77 & 73.25 & 82.00 & 73.62 \\

&\cellcolor{cyan!10}CoOp \cite{coop} & 64.50 & 75.53 & 75.00 & 77.68 & 57.50 & 70.70 & 47.50 & 49.50 & 61.13 & 68.35 \\

&\cellcolor{cyan!10}CoCoOp \cite{cocoop} & 47.50 & 51.68 & 76.50 & 68.63 & 58.50 & 57.28 & 60.00 & 47.59 & 60.63 & 56.30 \\

&\cellcolor{cyan!10}MaPLe \cite{maple} & 86.00 & 61.47 & 86.67 & 51.39 & 74.67 & 76.22 & 51.33 & 53.20 & 74.67 & 60.57 \\

&\cellcolor{cyan!10}LASP \cite{lasp} & 49.21 & 63.13 & 78.34 & 65.36 & 60.28 & 63.23 & 61.52 & 54.52 & 62.34 & 61.56 \\

&\cellcolor{cyan!10}PromptSRC \cite{promptsrc} & 87.33 & 63.28 & 87.17 & 65.06 & 67.60 & 67.56 & 52.30 & 54.35 & 73.60 & 62.56 \\

&\cellcolor{cyan!10}CLIPN \cite{clipn} & 88.64 & 66.21 & 88.35 & 70.32 & 73.24 & 71.02 & 59.28 & 60.14 & 77.38 & 66.92 \\

&\cellcolor{cyan!10}\textsc{StyLIP} \cite{stylip} & 89.18 & 68.93 & 89.84 & 74.27 & 76.69 & 71.58 & 65.15 & 61.66 & 80.22 & 69.11 \\

&\cellcolor{cyan!10}CLIPN + \textsc{StyLIP} & 88.67 & 70.48 & 88.39 & 80.32 & 85.34 & 77.40 & 83.97 & 76.50 & 86.59 & 76.18 \\

&\cellcolor{violet!20}MaPLe + SD & 88.73 & 78.50 & 85.60 & 78.47 & 80.60 & 79.80 & 80.22 & 80.43 & 83.79 & 79.30 \\

&\cellcolor{violet!20}PromptSRC + SD & 89.03 & 80.29 & 86.04 & 84.96 & 80.11 & 82.35 & 80.30 & 84.21 & 83.87 & 82.95 \\

&\cellcolor{violet!20}\textsc{StyLIP} + SD & 89.67 & 83.13 & 86.39 & 85.12 & 80.20 & 83.04 & 80.23 & 81.53 & 84.12 & 83.21 \\

\cmidrule(lr){2-12}

&\cellcolor{blue!20}\textbf{\textsc{ODG-CLIP}} &\cellcolor{blue!20}\cellcolor{blue!20}\textbf{97.55} &\cellcolor{blue!20}\textbf{94.50} &\cellcolor{blue!20}\textbf{96.40}&\cellcolor{blue!20}\textbf{95.60} &\cellcolor{blue!20}\textbf{95.33} &\cellcolor{blue!20}\textbf{95.45}&\cellcolor{blue!20}\textbf{93.44}&\cellcolor{blue!20}\textbf{92.35}&\cellcolor{blue!20}\textbf{95.68}&\cellcolor{blue!20}\textbf{94.48}\\

\bottomrule
\end{tabular}}
\label{tab_minidomainnet_open}
\end{center}
\end{table*}

\begin{table*}[htbp]
\caption{Comparative analysis for PACS, VLCS and Office-Home in closed-set setting over all the domain combinations following leave-one-domain-out protocol.}
\vspace*{-3mm}
\begin{center}
\scalebox{0.6607}{
\begin{tabular}{llccccc|ccccc|ccccc}
\toprule

&\multicolumn{1}{l}{\multirow{2}{*}{\textbf{Methods}}}&\multicolumn{5}{c}{\textbf{PACS}} &\multicolumn{5}{c}{\textbf{VLCS}}&\multicolumn{5}{c}{\textbf{Office-Home}}\\
\cmidrule(lr){3-7}\cmidrule(lr){8-12}\cmidrule(lr){13-17}
 &&\textbf{Art}&\textbf{Cartoon}&
 \textbf{Photo}&\textbf{Sketch}&
 \textbf{Avg}&\textbf{Caltech}&
 \textbf{LabelMe}&\textbf{Sun}&\textbf{P-VOC}&\textbf{Avg}&\textbf{Art}&\textbf{Clipart}&
 \textbf{Product}&\textbf{R-World}&
 \textbf{Avg}\\
\midrule
\multirow{3}{*}{\rotatebox{90}{CNN}} &SWAD \cite{swad} & 89.3 & 83.4 & 97.3 & 82.5 & 88.1 & 98.8 & 63.3 & 75.3 & 79.2 & 79.1 & 66.1 & 57.7 & 78.4 & 80.2 & 70.6 \\

&EoA \cite{arpit2022ensemble} & 90.5 & 83.4 & 98.0 & 82.5 & 88.6 & 99.1 & 63.1 & 75.9 & 78.3 & 79.1 & 69.1 & 59.8 & 79.5 & 81.5 & 72.5 \\
&DandelionNet \cite{dandelionnet} & 87.8 & 86.5 & 96.8 & 85.8 & 89.2 & 99.1 & 70.2 & 77.2 & 80.0 & 81.6 & 65.8 & 58.6 & 78.0 & 79.7 & 70.5 \\
\midrule

\multirow{7}{*}{\rotatebox{90}{CLIP-based}}&\cellcolor{grey}CLIP \cite{clip} & 96.21 & 98.07 & 98.65 & 86.62 & 94.89 & 98.73 & 69.05 & 82.56 & 78.23 & 82.14 & 74.58 & 67.94 & 84.85 & 86.21 & 78.40 \\

&\cellcolor{cyan!10}CoOp \cite{coop} & 97.85 & 98.64 & 99.70 & 92.23 & 97.11 & 98.58 & 70.20 & 84.28 & 80.31 & 83.34 & 77.32 & 72.10 & 88.43 & 87.46 & 81.33 \\

&\cellcolor{cyan!10}CoCoOp \cite{cocoop} & 97.42 & 98.18 & 99.54 & 91.02 & 96.54 & 98.93 & 73.18 & 85.21 & 82.76 & 85.02 & 77.45 & 72.03 & 87.92 & 86.81 & 81.05 \\

&\cellcolor{cyan!10}MaPLe \cite{maple} & 98.84 & 98.90 & 99.75 & 93.40 & 97.72 & 99.12 & 75.66 & 86.43 & 85.80 & 86.75 & 78.50 & 76.23 & 89.95 & 89.40 & 83.52 \\

&\cellcolor{cyan!10}LASP \cite{lasp} & 98.10 & 98.34 & 99.27 & 92.35 & 97.02 & 99.45 & 76.54 & 86.98 & 86.02 & 87.25 & 79.24 & 76.75 & 90.14 & 90.37 & 84.13 \\

&\cellcolor{cyan!10}PromptSRC \cite{promptsrc} & 98.79 & 99.02 & 99.50 & 94.76 & 98.02 & 99.61 & 75.30 & 85.39 & 85.07 & 86.34 & 78.97 & 75.82 & 90.31 & 90.44 & 83.89 \\

&\cellcolor{cyan!10}\textsc{StyLIP} \cite{stylip} & 98.73 & 99.15 & 99.97 & 94.82 & 98.17 & 99.70 & 75.84 & 87.08 & 86.22 & 87.21 & 81.54 & 78.78 & 91.67 & 91.75 & 85.94 \\

\cmidrule(lr){2-12}

&\cellcolor{blue!20}\textbf{\textsc{ODG-CLIP}} &\cellcolor{blue!20}\cellcolor{blue!20}\textbf{99.93} &\cellcolor{blue!20}\textbf{99.87} &\cellcolor{blue!20}\textbf{100.00}&\cellcolor{blue!20}\textbf{99.51} &\cellcolor{blue!20}\textbf{99.83} &\cellcolor{blue!20}\textbf{100.00}&\cellcolor{blue!20}\textbf{92.63}&\cellcolor{blue!20}\textbf{95.71}&\cellcolor{blue!20}\textbf{94.60}&\cellcolor{blue!20}\textbf{95.74} & \cellcolor{blue!20}\textbf{96.38} & \cellcolor{blue!20}\textbf{92.35} & \cellcolor{blue!20}\textbf{99.52} & \cellcolor{blue!20}\textbf{99.37} & \cellcolor{blue!20}\textbf{96.91}\\

\bottomrule
\end{tabular}}
\label{tab_pacsvlcsoh_closed}
\end{center}
\end{table*}

\begin{table*}[htbp]
\caption{Comparative analysis for Digits-DG and Mini-DomainNet in closed-set setting over all the domain combinations following leave-one-domain-out protocol.}
\vspace*{-3mm}
\begin{center}
\scalebox{0.6607}{
\begin{tabular}{llccccc|ccccc}
\toprule

&\multicolumn{1}{l}{\multirow{2}{*}{\textbf{Methods}}}&\multicolumn{5}{c}{\textbf{Digits-DG}} &\multicolumn{5}{c}{\textbf{Mini-DomainNet}}\\
\cmidrule(lr){3-7}\cmidrule(lr){8-12}
 &&\textbf{MNIST}&\textbf{MNIST-M}&
 \textbf{SVHN}&\textbf{SYN}&
 \textbf{Average}&\textbf{Clipart}&
 \textbf{Real}&\textbf{Painting}&\textbf{Sketch}&\textbf{Average}\\
\midrule

&\cellcolor{grey}CLIP \cite{clip} & 83.48 & 58.41 & 46.64 & 69.82 & 64.59 & 85.25 & 66.84 & 95.13 & 67.71 & 78.73 \\

&\cellcolor{cyan!10}CoOp \cite{coop} & 93.11 & 71.32 & 61.28 & 82.73 & 77.11 & 82.49 & 61.34 & 92.94 & 64.42 & 75.30 \\

&\cellcolor{cyan!10}CoCoOp \cite{cocoop} & 93.56 & 74.90 & 64.51 & 84.45 & 79.36 & 77.38 & 59.75 & 88.57 & 60.34 & 71.51 \\

&\cellcolor{cyan!10}MaPLe \cite{maple} & 94.25 & 75.68 & 66.72 & 84.33 & 80.25 & 81.27 & 62.58 & 88.29 & 63.32 & 73.87 \\

&\cellcolor{cyan!10}LASP \cite{lasp} & 95.87 & 75.61 & 65.91 & 82.28 & 79.92 & 80.51 & 58.30 & 85.14 & 58.72 & 70.67 \\

&\cellcolor{cyan!10}PromptSRC \cite{promptsrc} & 96.24 & 78.94 & 68.04 & 86.36 & 82.40 & 87.63 & 62.45 & 89.52 & 64.80 & 76.10 \\

&\cellcolor{cyan!10}\textsc{StyLIP} \cite{stylip} & 96.39 & 78.53 & 66.35 & 85.20 & 81.62 & 89.36 & 67.63 & 94.57 & 70.14 & 80.43 \\

\cmidrule(lr){2-12}

&\cellcolor{blue!20}\textbf{\textsc{ODG-CLIP}} &\cellcolor{blue!20}\cellcolor{blue!20}\textbf{99.48} &\cellcolor{blue!20}\textbf{96.38} &\cellcolor{blue!20}\textbf{91.22}&\cellcolor{blue!20}\textbf{98.42} &\cellcolor{blue!20}\textbf{96.38} &\cellcolor{blue!20}\textbf{98.54}&\cellcolor{blue!20}\textbf{92.37}&\cellcolor{blue!20}\textbf{99.42}&\cellcolor{blue!20}\textbf{96.25}&\cellcolor{blue!20}\textbf{96.65}\\

\bottomrule
\end{tabular}}
\label{tab_digitsmini_closed}
\end{center}
\end{table*}

\end{document}